%% file: neurips_2026.tex
\documentclass{article}

    \PassOptionsToPackage{numbers, sort&compress}{natbib}

\usepackage[preprint]{neurips_2026}

\usepackage[utf8]{inputenc} 
\usepackage[T1]{fontenc}    
\usepackage{hyperref}       
\usepackage{url}            
\usepackage{booktabs}       
\usepackage{amsfonts}       
\usepackage{nicefrac}       
\usepackage{microtype}      
\usepackage{xcolor}         

\definecolor{skyblue}{RGB}{80, 160, 200}
\definecolor{darkgreen}{RGB}{0,90,60}
\usepackage{amsmath}
\usepackage{graphicx}
\usepackage{caption}
\usepackage{cleveref}
\usepackage{multirow}
\usepackage{amssymb}
\usepackage{subcaption}

\usepackage{tabularx}
\newcolumntype{Y}{>{\centering\arraybackslash}X}

\crefname{figure}{Fig.}{Figs.}
\crefname{table}{Tab.}{Tabs.}
\crefname{section}{Sec.}{Secs.}

\Crefname{figure}{Figure}{Figures}
\Crefname{table}{Table}{Tables}
\Crefname{section}{Section}{Sections}

\title{Partially Observable Zero-shot coordination by Predicting Intention of Partner}

\author{
\parbox{\textwidth}{
\centering
Jinnyeong Yang\thanks{Equal contribution.} \quad
Yuhwan Jeong\footnotemark[1] \quad
Hoyong Kwon\footnotemark[1] \quad
Minseok Kim\footnotemark[2] \quad
Jihun Kim\footnotemark[2] \\
Kuk-Jin Yoon \\
KAIST, Visual Intelligence Lab \\
\texttt{\{jinnyeong6118, jeongyh98, kwonhoyong3, alstjrx1x1, jihun1998,}\\
\texttt{kjyoon\}@kaist.ac.kr}
}
}
\vspace{-10pt}

\begin{document}

\maketitle

\input{sec/0_abstract}
\input{sec/1_intro_new}
\input{sec/2_related_works}
\input{sec/3_Preliminaries}
\input{sec/4_method}
\input{sec/5_Experiments}
\input{sec/6_conclusion}


{
\small
\bibliography{bib}
}

\clearpage
\input{sec/X_supple}

\end{document}

%% file: sec/0_abstract.tex
\begin{abstract}
Zero-shot coordination in embodied settings requires acting while the partner is intermittently out of view, leaving existing methods with ambiguous partner representations and uncertainty over hidden partner states. We propose Predicting Intention of Partner (PIP) to jointly address these challenges. PIP uses a Joint-view VAE to distill richer training-time evidence from the union of both agents' local observations into a partner representation available from local observations alone. Partner-state Belief networks further infer the partner's hidden location and behavioral tendencies from the ego agent's interaction history. We evaluate PIP in Burrito-PO, Overcooked-PO, and a Melting Pot substrate, together with a human evaluation in Burrito-PO. PIP attains the highest mean performance among the compared methods across all three benchmarks. Human evaluation and diagnostic analyses further support coordination with unseen partners and the contributions of both components under partner occlusion.
\end{abstract}
\vspace{-10pt}

%% file: sec/1_intro_new.tex
\section{Introduction}
\label{sec:intro}
As AI agents and robots are increasingly deployed in real-world interactive systems, human-AI collaboration has become a central goal of artificial intelligence. 
In such settings, an agent must coordinate with diverse unseen humans or partner agents that were never encountered during training. This requirement has motivated the study of zero-shot coordination (ZSC)~\citep{jha2025crossenvironment, lupu2021trajectory, li2023cooperative}, whose central difficulty is the shift of partner distribution between training and deployment. Agents trained through self-play~\citep{hu2020other, sarkar2023diverse} often converge to arbitrary conventions and fail when paired with partners that follow different ones. To mitigate this, prior work has expanded the diversity of the partner population used during training~\citep{charakorn2023generating, strouse2021fcp, zhao2023maximum}. Representing the space of possible partners with a finite set of policies, however, is inherently limited. Generative prior-based methods~\citep{chaudhary2026goat, li2025talent, liang2024gamma} have therefore emerged as a promising alternative. By modeling a continuous latent space of partner behaviors from coordination trajectories, they expose the target policy to partners beyond the observed population, and have achieved remarkable results in fully observable environments such as Overcooked~\citep{carroll2019utility}.


%

\begin{figure}[h]
    \centering
    \includegraphics[width=.60\linewidth]{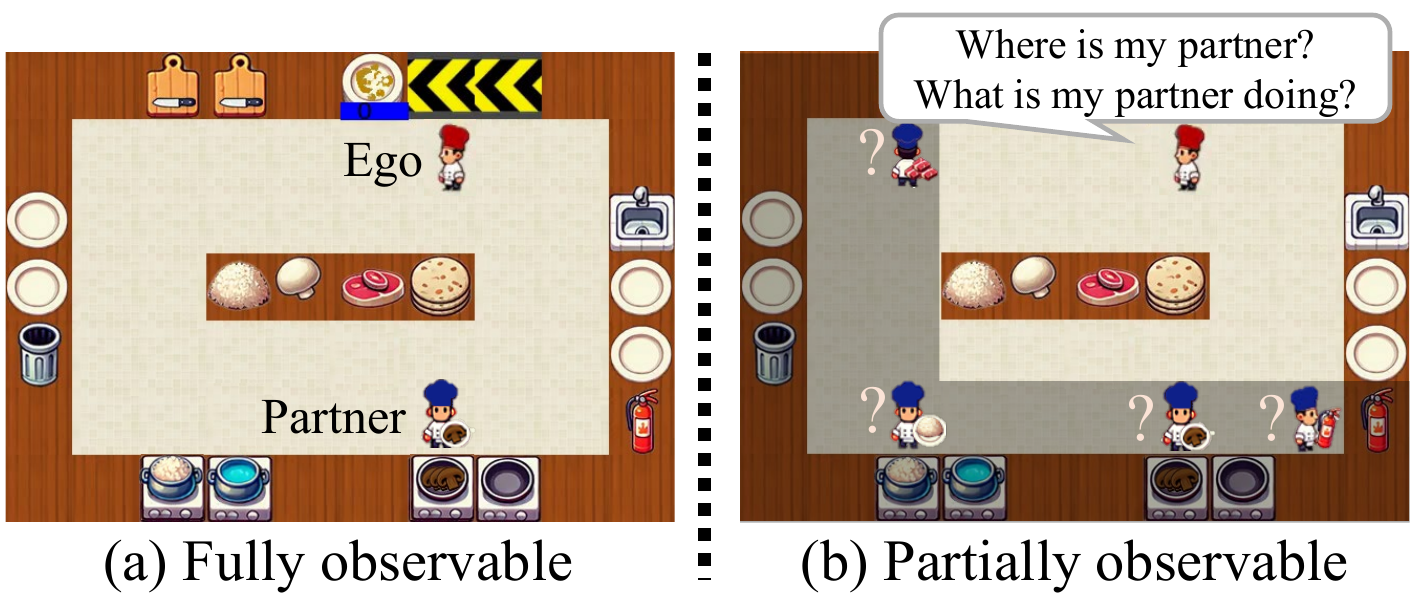}
    \caption{Illustration of coordination under dynamic and spatial partial observability in Burrito-PO. (a) Under full observation, the partner state is directly available. (b) Under a restricted local view, the partner may leave the visible region, creating ambiguity over its location and behavior.}
    \label{fig:teaser}
    \vspace{-10pt}
\end{figure}

In practical collaboration, however, limited sensing and the movement of agents rarely afford full observability over either the task state or the partner~\citep{omidshafiei2017deep, ong2010planning}.
Under such partial observability (PO), coordination cues are confined to the ego agent's observable region, as illustrated in~\cref{fig:teaser}.
The agent must coordinate with unseen partners from incomplete information about partner behavior and task state. Within ZSC, partial observability has been studied primarily in the informational sense exemplified by Hanabi~\citep{bard2020hanabi}, where the hidden information is prescribed by the game rules. In such settings the partner's actions remain visible, and what is concealed is a fixed set of variables. Embodied collaboration instead induces spatial and dynamic partial observability. 

What the ego agent perceives changes over time with its restricted field of view and with the movement of both agents. The information asymmetry here is therefore induced by viewpoint.
Settings of this kind have only recently been taken up as ZSC problems, with existing studies focusing on environment design~\citep{gessler2025overcookedv2} or symmetry~\citep{muglich2025expected}.
Yet the failure modes induced by viewpoint-based partial observability itself remain underexplored.

%
In this work, we study ZSC under dynamic and spatial partial observability. We identify two key challenges that arise when VAE-based generative ZSC is transferred to partially observable environments. First, partial observability degrades the partner diversity captured in the generator's latent space. Because informative segments of the partner's motion are occluded outside the field of view, the encoder receives distinct partner trajectories as similar local observations. Behavior patterns that are separable under richer observations thus collapse into overlapping latent regions, restricting the coverage of the modeled partner space. To address this, we introduce a Joint-view VAE. 
A teacher encoder receives the union of the two agents' local observations, and its posterior is distilled into a student encoder that sees only the ego-centric observation.
This preserves behaviorally meaningful variation under limited input and retains diversity in the generated partners. 
Here, the joint view denotes additional partial-view evidence available only during VAE training, not access to the unobservable global state. The resulting generator therefore remains deployment-compatible.

%
Second, partial observability restricts the evidence on which the target policy bases its decisions. Once the partner leaves the ego agent's observable region, essential information such as its current location, recent action, and intended subtask is lost~\citep{foerster2019bayesian, gu2022online, kolb2024inferring, nguyen2023social, rahman2023general}. Since coordination decisions depend heavily on where the partner is and what it will do next, the policy must infer hidden partner-related information from its own observation history. Inspired by belief-based modeling in partially observable reinforcement learning~\citep{igl2018deep, kaelbling1998planning, Zintgraf2020VariBAD}, we introduce Partner-state Belief networks. A Location Belief estimates the position of the partner while it is out of view, and a Behavioral Belief predicts what the partner is likely to do. Together they allow the policy to compensate for uncertainty about the partner and to act in a belief-aware manner. 
Importantly, we follow a decentralized PO protocol during policy execution. Logged partner information is used only as auxiliary supervision for the belief networks at the PPO update stage, and is never provided as input to the policy. At deployment, the policy therefore acts solely from its own observations and its self-estimated beliefs.

%

Integrating these two components, we propose Predicting Intention of Partner (PIP), a framework for zero-shot coordination under dynamic and spatial partial observability. The Joint-view VAE improves the diversity of generated partners, and the Partner-state Belief enables belief-aware decision making. PIP thus jointly addresses partner generation and policy learning from incomplete observations.
We evaluate PIP on Overcooked-PO, Burrito-PO, and Melting Pot~\citep{carroll2019utility, li2025talent, agapiou2022melting}, benchmarks that combine dynamic cooperative tasks with spatial partial observability. PIP achieves strong coordination performance with both unseen partners and humans, and further analyses demonstrate that each component operates as intended and contributes to robust coordination.
Our main contributions are as follows:
\begin{itemize}
\item To the best of our knowledge, our work is the first to address the failure modes that dynamic spatial partial observability induces in both partner generation and policy learning in ZSC.
\item We propose PIP, which combines a Joint-view VAE for diverse partner generation with Partner-state Belief networks to handle information asymmetry in spatial environment. 
\item Extensive experiments including human study on the Burrito-PO, Overcooked-PO, and Melting Pot benchmarks demonstrate the effectiveness of PIP.
\end{itemize}

%% file: sec/2_related_works.tex
\section{Related works}
\label{sec:related_works}

\paragraph{Zero-Shot Coordination and Partner Diversity.}
Zero-shot coordination requires agents to collaborate with previously unseen partners, including independently trained agents and humans~\citep{stone2010ad,hu2020other}. Other-Play~\citep{hu2020other} and Any-Play~\citep{lucas2022any} reduce reliance on arbitrary coordination conventions. CEC~\citep{jha2025crossenvironment} promotes coordination between agents trained in different environments. Another major direction is population-based training, which exposes a target policy to diverse partners and conventions. Existing methods construct partner populations or explicitly optimize their diversity~\citep{strouse2021fcp,zhao2023maximum,rahman2023generating,lupu2021trajectory}. GAMMA~\citep{liang2024gamma}, TALENTS~\citep{li2025talent}, and GOAT~\citep{chaudhary2026goat} use generative latent spaces to produce diverse or challenging partners. E3T~\citep{yan2023e3t} avoids a pretrained population by constructing partners from mixtures of the ego and random policies while jointly learning partner action prediction. Collectively, these approaches improve generalization by exposing the target policy to diverse behaviors and coordination conventions during training.

\paragraph{Zero-Shot Coordination under Partial Observability.}
PO-ZSC includes different settings depending on which information is unobserved and how it becomes unavailable. A representative line of work considers environments such as Hanabi~\citep{bard2020hanabi}, where private or asymmetric information is defined by the game structure. In Hanabi, agents have access to different card information, and actions serve both task execution and communication. Prior studies combine compatibility with novel partners and belief-based inference of hidden information~\citep{hu2020other,lupu2021trajectory,foerster2019bayesian,muglich2022generalized}. ERS~\citep{muglich2025expected} instead broadens the symmetry class to discourage incompatible conventions in Dec-POMDPs. These studies primarily address information asymmetry imposed by game rules. In spatial coordination environments, however, observability changes with agent positions and movements. OvercookedV2~\citep{gessler2025overcookedv2} introduces restricted fields of view and asymmetric task information and evaluates existing ZSC methods under spatial partial observability.
However, existing methods handle the resulting intermittent loss of partner observations only implicitly.
To the best of our knowledge, prior work that jointly addresses incomplete partner representations and explicitly infers the positions and actions of currently unobserved partners remains limited.

%% file: sec/3_Preliminaries.tex
\section{Preliminaries}

\paragraph{Two-Player Partially Observable Markov Decision Process.}

A two-player partially observable Markov decision process extends the standard two-player Markov decision process to settings with incomplete information. Formally, the environment is defined as $\mathcal{M} = (\mathcal{S}, \mathcal{A}^0, \mathcal{A}^1, \mathcal{O}^0, \mathcal{O}^1, P, R, \Omega, \gamma)$, where $\mathcal{S}$ denotes the global state space, and $\mathcal{A}^i$ and $\mathcal{O}^i$ denote the action and observation spaces of agent $i \in \{0,1\}$, respectively. $P(s' \mid s, a^0, a^1)$ is the transition dynamics, $R(s, a^0, a^1, s')$ is the shared reward function, $\Omega(o^0, o^1 \mid s)$ denotes the observation function, and $\gamma \in [0,1)$ is the discount factor. At each timestep $t$, both agents receive local observations $(o_t^0,o_t^1) \sim \Omega(\cdot \mid s_t)$ rather than the full global state $s_t$, and each agent selects an action $a_t^i \sim \pi_i(\cdot \mid h_t^i)$ conditioned on its local action-observation history $h_t^i = (o_0^i, a_0^i, \dots, o_t^i)$. The joint transition then follows $s_{t+1} \sim P(\cdot \mid s_t, a_t^0, a_t^1)$. The objective is to learn policies $(\pi_0, \pi_1)$ that maximize the expected discounted return
$\max\limits_{\pi_0,\pi_1} \mathbb{E}\left[\sum_{t=0}^{H} \gamma^t R(s_t,a_t^0,a_t^1,s_{t+1})\right]$. In this work, the partner's strategy is not directly observed and must be inferred from local interaction history alone.

\paragraph{Generative Latent-Variable ZSC.}
Recent generative latent-variable approaches to ZSC~\citep{liang2024gamma,chaudhary2026goat} represent a cooperation strategy as a latent variable $z \in \mathcal{Z}$ in a continuous latent space. These methods first collect partner trajectory data, such as trajectories generated by MEP~\citep{zhao2023maximum}-trained policies. They then learn a generative representation of partner behaviors to model diverse coordination strategies. Let $\tau=(o^p_0,a^p_0,\dots,o^p_T,a^p_T)$ denote a trajectory generated by a partner policy, and let $h^p_t=(o^p_0,a^p_0,\dots,o^p_t)$ be the partner's history up to time $t$. An encoder $q_\phi(z \mid \tau)$ maps the trajectory to a posterior over latent strategies, where $z$ captures high-level behavioral factors such as convention, preference, or execution style. Conditioned on $z$, a decoder $p_\theta(a^p_t \mid z,h^p_t)$ predicts the partner's next action, thereby defining a latent-conditioned partner policy $\mu_z(a^p_t \mid h^p_t) := p_\theta(a^p_t \mid z,h^p_t)$. The generative model is trained with a variational objective
\begin{equation}
\begin{split}
\mathcal{L}_{\mathrm{gen}}(\theta,\phi)
=
\mathbb{E}_{\tau \sim \mathcal{D}}
\biggl[
\mathbb{E}_{z \sim q_\phi(z \mid \tau)}
\Bigl[
\sum_{t=0}^{T} \log p_\theta(a^p_t \mid z,h^p_t)
\Bigr] 
-
\beta\,D_{\mathrm{KL}}\!\left(q_\phi(z \mid \tau)\,\|\,p(z)\right)
\biggr],
\end{split}
\end{equation}
where $p(z)$ is typically chosen as a simple prior such as a standard Gaussian. After training, latent-conditioned partner policies are instantiated by sampling $z$ and using the learned decoder.

\paragraph{Partial-Observation Protocol.}
We follow the centralized training with decentralized execution (CTDE) paradigm for multi-agent reinforcement learning. To consider a more realistic partial observability setting, the policy is trained from PO rollouts collected by the agents, without using unobserved global state as policy input. During PPO training, logged partner-state information from these rollouts is used only as auxiliary supervision for the belief heads, and is not provided to the actor as input or used during execution. Thus, throughout decentralized execution, the policy is conditioned only on the agent's own ego-centric observation history.

%% file: sec/4_method.tex
\section{Proposed method}
\label{sec:method}

Our method, PIP, consists of two components for PO-ZSC. The Joint-view VAE improves local-view partner-behavior modeling with richer training-time evidence. The Partner-state Belief policy maintains beliefs over the partner's behavior and coarse location. Together, they address partner generation and policy learning under hidden partner states.

\subsection{Joint-view VAE}
\label{sec:pvvae}

Building on the generative latent-variable ZSC formulation, we consider the PO case where the behavior encoder receives only an agent-centered observation-action sequence. Under strict partial observability, an agent's local view may omit key interaction cues, such as the partner's location.

When these cues are missing, the same local trajectory can be consistent with multiple partner behaviors. This ambiguity makes it difficult for the encoder to infer a clear latent strategy from partial observations.
As a result, distinct interaction modes can be mapped to nearby or overlapping regions in the latent space, making the learned decoder less effective as a behavior generator. To address this issue, we propose a \textbf{Joint-view VAE (JV-VAE)}, which uses a teacher posterior from richer evidence to guide a student posterior conditioned on the agent-centered partial observation.

Let $i \in \{0,1\}$ denote the agent whose behavior is being encoded, and let $\bar{i}=1-i$ denote the other agent. The teacher is given a wider view of the interaction by combining the two agents' local observations, while the student is kept under the same input constraint as the vanilla VAE. We define
\begin{equation}
\begin{gathered}
  u_t^{i} = \mathrm{Union}(o_t^{i}, o_t^{\bar{i}}), \qquad
  x_{T,t}^{i} = \mathrm{Concat}(u_t^{i}, o_t^{i}), \qquad
  x_{S,t}^{i} = o_t^{i}.
\end{gathered}
\end{equation}
The teacher-side input $x_{T,t}^{i}$ uses both the union view and $o_t^i$, whereas the student-side input $x_{S,t}^{i}$ uses only $o_t^i$. The union view contains cells visible to either agent, while cells invisible to both agents remain masked. Thus, the teacher's joint input remains partial rather than a full observation and does not use the unobserved global state.

This joint evidence helps the teacher posterior distinguish behavior modes that are ambiguous from the student input alone. Matching this posterior encourages the student to preserve meaningful diversity in the latent space while keeping the PO input interface unchanged.

For each branch $\bullet \in \{T,S\}$, let $\tau_\bullet^i = \{(x_{\bullet,t}^{i}, a_t^i)\}_{t=1}^{H}$ denote the input-action trajectory used to infer the latent variable. The encoder is recurrent over this trajectory, so the latent is inferred from a temporal summary rather than from a single frame. We parameterize the posterior as
\begin{equation}\label{eq-pv-vae-posterior}
  q_\bullet(z \mid \tau_\bullet^i) = \mathcal{N}\!\left(\mu_\bullet(\tau_\bullet^i), \mathrm{diag}(\sigma_\bullet^{2}(\tau_\bullet^i))\right).
\end{equation}

Although downstream target policy training does not use the encoder, the posterior still shapes how behavior trajectories are organized in the latent space during VAE training. In this sense, posterior distillation improves the learned behavior representation by encouraging trajectories with different interaction modes to be better separated in the latent space.

Given these branch-specific posteriors, training follows a two-stage distillation procedure. We first train the teacher as a VAE on the richer partial-view input. We then freeze the teacher and train the student with its own negative ELBO together with a posterior distillation term,
\begin{equation}\label{eq-pv-vae-kd}
\begin{gathered}
  \mathcal{L}_{S} = \mathcal{L}_{S}^{\mathrm{VAE}} + \lambda_{\mathrm{KD}}\,\mathcal{L}_{\mathrm{KD}}, \qquad
  \mathcal{L}_{\mathrm{KD}} = \mathbb{E}_{i \in \{0,1\}}\!\left[\mathrm{KL}\!\left(q_T(z \mid \tau_T^i) \,\|\, q_S(z \mid \tau_S^i)\right)\right].
\end{gathered}
\end{equation}
Here, $\mathcal{L}_{S}^{\mathrm{VAE}}$ denotes the student's negative ELBO.

In subsequent policy learning, we use the trained student decoder as a deployment-compatible partner generator for target-policy training.

\subsection{Partner-State Belief Policy}
\label{sec:psbp}

Although the Joint-view VAE improves the generative partner model, it alone does not resolve the target agent’s decision-making problem under partial observability. At deployment, the agent must select ego actions from its ego-centric history, where the partner’s current location, intention, and action tendency may be hidden.
While the recurrent actor can implicitly summarize the interaction history, we use a separately supervised belief state to explicitly retain partner-related information that may become unavailable in the current observation.
We therefore augment the target policy with a \textbf{Partner-state Belief} module updated from the agent’s local observation history during execution.

The main design principle is to decouple the action state from the belief state used for partner-state estimation. Let $o_t$ denote the ego-centric partial observation. The actor updates its action state $h_t^{\mathrm{act}} = f_{\mathrm{act}}(e_t, h_{t-1}^{\mathrm{act}})$ and uses it to select the ego action, where $e_t$ is the encoded feature of $o_t$. In parallel, the belief module updates a separate belief state $h_t^{\mathrm{bel}} = f_{\mathrm{bel}}(\tilde{e}_t, h_{t-1}^{\mathrm{bel}})$, where $\tilde{e}_t$ is obtained by applying the same observation encoder to a masked observation $\tilde{o}_t$. Here, $\tilde{o}_t$ is constructed by masking the agent-orientation channel in $o_t$. This masking removes a direct orientation shortcut and encourages the belief module to infer partner state from the remaining partial observation history.

\paragraph{Behavioral Belief.}
The first auxiliary objective learns the Behavioral Belief over future partner actions. Motivated by theory-of-mind accounts~\citep{rabinowitz2018machine, 10.5555/3535850.3535957,wu2021too} that distinguish a partner's stable behavioral character from their situation-specific reaction, we separate Behavioral Belief into reaction and tendency components. We use two prediction heads, $\ell_t^{\mathrm{rec}} = \psi_{\mathrm{rec}}(h_t^{\mathrm{bel}})$ and $\ell_t^{\mathrm{ten}} = \psi_{\mathrm{ten}}(h_t^{\mathrm{bel}})$, where $\mathrm{rec}$ and $\mathrm{ten}$ denote reaction and tendency, respectively, and $\psi_{\mathrm{rec}}$ and $\psi_{\mathrm{ten}}$ map the belief state to behavior logits. Given the partner action sequence $a_t^p$, the corresponding soft targets are
\begin{equation}
\begin{gathered}
q_t^{\mathrm{rec}} = \mathcal{N}\left(\sum_{k=1}^{K_{\mathrm{off}}-1} \gamma_{\mathrm{rec}}^{k-1} \mathbf{1}[a_{t+k}^{p}]\right), \qquad
q_t^{\mathrm{ten}} = \mathcal{N}\left(\sum_{k=K_{\mathrm{off}}}^{T-t} \gamma_{\mathrm{ten}}^{k-K_{\mathrm{off}}} \mathbf{1}[a_{t+k}^{p}]\right).
\end{gathered}
\end{equation}
Here, $\mathcal{N}(\cdot)$ normalizes a nonnegative vector to the probability simplex. The reaction target emphasizes short-term partner actions, whereas the tendency target captures more persistent behavioral patterns. Both heads are trained with soft cross-entropy losses $\mathcal{L}_{\mathrm{rec}}$ and $\mathcal{L}_{\mathrm{ten}}$.

\paragraph{Location Belief.}
The second auxiliary objective learns the Location Belief. For each offset $\Delta \in \mathcal{D}$, where $\mathcal{D}$ includes the current offset $\Delta=0$ and multiple future offsets, the location head predicts $\ell_{t,\Delta}^{\mathrm{loc}} = \psi_{\mathrm{loc}}(h_t^{\mathrm{bel}}, e_\Delta)$, where $e_\Delta$ is a learned offset embedding. The target $\tilde{b}_{t,\Delta}$ is a soft distribution over coarse cells that reflects both the partner's future position and the uncertainty caused by partial observability. We construct it by mixing two coarse-cell distributions, a Gaussian future-cell target $g_{t,\Delta}^{\mathrm{cell}}$ and a reachability prior $g_{t,\Delta}^{\mathrm{reach}}$ from the last seen partner location, using a coefficient $\alpha \in [0,1]$,
\begin{equation}
\tilde{b}_{t,\Delta} = \alpha g_{t,\Delta}^{\mathrm{cell}} + (1-\alpha) g_{t,\Delta}^{\mathrm{reach}}.
\end{equation}
The future-cell target is centered at the true future partner cell and masked to physically reachable cells. The reachability prior assigns probability mass to coarse cells whose distance from the last seen partner cell is no larger than the elapsed time since the partner was last observed. If the partner has not been observed for more than $H_{\max}$ steps, we replace the reachability prior with a uniform spatial prior. This mixture avoids treating the hidden partner location as fully observed when the partner has been occluded.
The location head is trained with a normalized weighted KL loss, where $w_\Delta$ controls the relative contribution of each offset,
\begin{equation}
\mathcal{L}_{\mathrm{loc}} = \frac{\sum_{\Delta} w_\Delta \, \mathrm{KL}\!\left(\mathrm{softmax}(\ell_{t,\Delta}^{\mathrm{loc}})\,\|\,\tilde{b}_{t,\Delta}+\epsilon\right)}{\sum_{\Delta} w_\Delta}.
\end{equation}
Within each offset, samples in which the partner is unobserved receive larger weights, since Location Belief is most useful under partial observability.

The Partner-state Belief is then passed to the actor through a stable additive interface. We first project the belief state as $r_t = \phi_{\mathrm{bel}}(\mathrm{stopgrad}(h_t^{\mathrm{bel}}))$, where $\phi_{\mathrm{bel}}$ is a projection layer. The stop-gradient operation prevents PPO gradients from directly reshaping the belief state and keeps the belief representation primarily governed by the Behavioral Belief and Location Belief objectives. We also compute a normalized uncertainty feature from the entropy of the current Location Belief, $u_t = H(\mathrm{softmax}(\ell_{t,0}^{\mathrm{loc}}))/\log N_{\mathrm{cell}}$, and define the actor-side belief message as $m_t^{\mathrm{bel}}=[r_t,u_t]$.

The actor receives this belief message as an additive update to its action state, $\Delta h_t^{\mathrm{act}} = W_{\mathrm{bel}}m_t^{\mathrm{bel}} + c_{\mathrm{bel}}$, where $W_{\mathrm{bel}}$ and $c_{\mathrm{bel}}$ denote the linear projection and bias of the belief-injection layer, respectively. The actor then uses $\bar{h}_t^{\mathrm{act}} = h_t^{\mathrm{act}} + \Delta h_t^{\mathrm{act}}$ for action selection. The parameters $W_{\mathrm{bel}}$ and $c_{\mathrm{bel}}$ are initialized to zero, so the initial policy is identical to the recurrent baseline.

We train the belief-augmented actor with PPO from environment rewards and supervise the belief module with training-only partner-state targets. We add a lightweight alignment regularizer $\mathcal{L}_{\mathrm{aux}}$ on the belief-to-actor interface, blocking gradients to the belief RNN. The combined loss is
\begin{multline}
\mathcal{L}_{\mathrm{train}} = \mathcal{L}_{\mathrm{PPO}}\bigl(\pi(\cdot \mid \bar{h}_t^{\mathrm{act}})\bigr) 
+ \lambda_{\mathrm{behavior}}\left(\alpha_{\mathrm{rec}}\mathcal{L}_{\mathrm{rec}} + \alpha_{\mathrm{ten}}\mathcal{L}_{\mathrm{ten}}\right)
+ \lambda_{\mathrm{loc}}\mathcal{L}_{\mathrm{loc}}
+ \lambda_{\mathrm{aux}}\mathcal{L}_{\mathrm{aux}}.
\end{multline}
Since the injected belief message is detached from the belief state, PPO gradients do not directly update the belief RNN. The belief RNN is trained by the Behavioral Belief and Location Belief losses, while PPO learns how to use the injected belief message for reward maximization.

%% file: sec/5_Experiments.tex
\section{Experiments}
\label{sec:exp}

 We independently instantiate and train all methods for every benchmark environment, including each layout. For every environment, we separately train the partner population, construct the trajectory dataset, train the generative partner model, and optimize the target policy. No policy, VAE, or partner population is transferred across environments. Across all benchmarks, deployed policies and evaluation partners operate only from local observations. The joint-view teacher is used solely during VAE training and uses the union of both agents' local observations, while the distilled student and all deployed policies remain restricted to local observations.

We compare PIP with the generative-partner baselines GAMMA~\citep{liang2024gamma} and GOAT~\citep{chaudhary2026goat} in all three benchmarks. For Burrito and Overcooked, we restrict fully observable environments to egocentric local views. Burrito-PO additionally includes FCP~\citep{strouse2021fcp}, MEP~\citep{zhao2023maximum}, E3T~\citep{yan2023e3t}, and ERS~\citep{muglich2025expected}. Performance is measured using the task reward obtained with partners excluded from target-policy training. Environment-specific architectures, hyperparameters, partner construction procedures, and additional implementation details are provided in the \textit{appendices}.

\begin{table}[t]
\centering
\small
\setlength{\tabcolsep}{1mm}
\caption{Evaluation with 12 held-out behavior-preference partners unseen during training. The best result in each layout is shown in \textbf{bold}. Mean $\pm$ standard error.}
\begin{tabular}{lcccc}
\toprule
Method & Open & Hallway & FC & Ring \\
\midrule
FCP & 141.6 $\pm$ 9.6 & 57.1 $\pm$ 2.7 & 21.9 $\pm$ 6.0 & 48.3 $\pm$ 6.1 \\
MEP & 115.9 $\pm$ 10.3 & 58.2 $\pm$ 2.6 & 17.6 $\pm$ 5.8 & 51.1 $\pm$ 5.4 \\
E3T & 55.3 $\pm$ 3.6 & 44.5 $\pm$ 5.7 & 21.0 $\pm$ 4.1 & 27.5 $\pm$ 3.1 \\
ERS & 101.0 $\pm$ 7.8 & 36.6 $\pm$ 4.5 & 24.0 $\pm$ 4.3 & 48.9 $\pm$ 5.5 \\
GAMMA & 163.3 $\pm$ 10.6 & 57.7 $\pm$ 3.3 & 19.8 $\pm$ 5.3 & 48.5 $\pm$ 5.6 \\
GOAT & 165.5 $\pm$ 10.6 & 61.5 $\pm$ 5.1 & 19.0 $\pm$ 5.3 & 77.3 $\pm$ 5.9 \\
PIP (Ours) & \textbf{249.0 $\pm$ 12.2} & \textbf{78.1 $\pm$ 5.3} & \textbf{25.4 $\pm$ 8.9} & \textbf{125.8 $\pm$ 5.6} \\
\bottomrule 
\end{tabular}
\label{tab:main_result}
\end{table}

\subsection{Burrito-PO}
\label{sec:burrito_po}
Burrito-PO is our primary benchmark for zero-shot coordination under ego-centric partial observability. It adapts the cooperative Burrito environment~\citep{li2025talent} by restricting each agent to a $7 \times 7$ window centered on itself and masking cells outside this window. We consider four layouts, Open, Hallway, Forced-Coord (FC), and Ring. Following prior work~\citep{wang2024zsc,li2025talent}, we construct 24 candidate partners per layout and select 12 behaviorally diverse partners for evaluation. For each layout, every selected partner is evaluated over 10 episodes. 
We report mean performance and standard error.


\Cref{tab:main_result} reports ZSC performance against held-out partners under partial observability. PIP achieves the best performance across all four layouts. The relative improvement is particularly large on Ring, where the movable region forms a narrow loop-like route. The partner's future movement is strongly constrained by this structure, allowing the belief module to anticipate the partner's hidden position and movement direction more reliably from local observations. This result suggests that Partner-state Belief is particularly beneficial when the partner is frequently outside the ego agent's field of view and the policy must infer hidden partner states to anticipate coordination opportunities.


\subsection{Overcooked-PO}
\label{sec:overcooked_po}


Overcooked-PO is a local-observation variant of Overcooked~\citep{carroll2019utility} that retains the original transition dynamics, six primitive actions, task objective, and reward structure. We evaluate on CounterCircuit using a $5 \times 5$ ego-centered local observation. Consequently, the partner and its current subtask may become unobservable when the agents move across different regions of the kitchen. All methods are evaluated with the same 12 behavior-preference partners.


\begin{table}[t]
\centering
\begin{minipage}[t]{0.42\linewidth}
\small
\captionof{table}{Performance in two additional coordination environments, Overcooked-PO and Melting Pot, respectively.}
\begin{tabular}{lcc}
\toprule
Method & Overcooked-PO & Melting Pot \\
\midrule

GAMMA & 349.6 $\pm$ 5.14 & 171.3 $\pm$ 5.2 \\
GOAT & 325.9 $\pm$ 5.73 & 153.1 $\pm$ 5.2 \\
PIP (Ours) & \textbf{402.34 $\pm$ 3.82} & \textbf{213.5 $\pm$ 11.2} \\
\bottomrule
\end{tabular}
\label{tab:additional_env_results}
\end{minipage}
\hfill
\begin{minipage}[t]{0.53\linewidth}
    \centering
    \small
    \captionof{table}{Ablation study in the Open layout. Behavioral and Location denote the Behavioral Belief and Location Belief modules, respectively.}
    \begin{tabular}{ccccc}
    \toprule
    & JV-VAE & Behavioral & Location & Reward \\
    \midrule
    (a) &              &              &              & 163.3 $\pm$ 10.6 \\
    (b) & $\checkmark$ &              &              & 181.6 $\pm$ 9.1 \\
    (c) & $\checkmark$ & $\checkmark$ &              & 211.0 $\pm$ 8.5 \\
    (d) & $\checkmark$ &              & $\checkmark$ & 213.2 $\pm$ 5.2 \\
    (e) &              & $\checkmark$ & $\checkmark$ & 183.1 $\pm$ 12.3 \\
    (f) & $\checkmark$ & $\checkmark$ & $\checkmark$ & 249.0 $\pm$ 12.2 \\
    \bottomrule
    \end{tabular}
    \label{tab:ablation_main}
\end{minipage}

\end{table}

As shown in~\cref{tab:additional_env_results}, PIP achieves the best performance among the compared methods on CounterCircuit. This result suggests that PIP remains effective in another $5\times5$ grid-based cooking environment under local observations.

\subsection{Melting Pot}
\label{sec:melting_pot}


 We further evaluate PIP in Melting Pot's~\citep{agapiou2022melting} \texttt{coop\_mining} substrate, instantiated as a two-player configuration. 
 This benchmark differs from the cooking environments through its native $88 \times 88$ ego-centric RGB observation, eight primitive actions, and cooperative mining objective. Each target policy is evaluated with three official bot partners that are excluded from target-policy training.



As shown in~\cref{tab:additional_env_results}, PIP achieves the best performance among the compared methods in Melting Pot. Together with the results on Burrito-PO and Overcooked-PO, this finding suggests that PIP remains effective across observation representations ranging from structured local grids to native ego-centric RGB inputs, as well as with primitive control and a cooperative objective distinct from the cooking tasks. However, the use of official bots precludes direct cross-benchmark comparisons.

\subsection{Human Evaluation}
\label{sec:human_eval}

We recruit 66 participants and ask each participant to interact with three anonymized partner agents corresponding to GAMMA, GOAT, and PIP in Burrito-PO. The agent order is randomized across participants. After each interaction, participants rate coordination quality via a survey. After all three interactions, they rank each agent by ease of cooperation.

\begin{figure}[t]
\centering
\includegraphics[width=.99\linewidth]{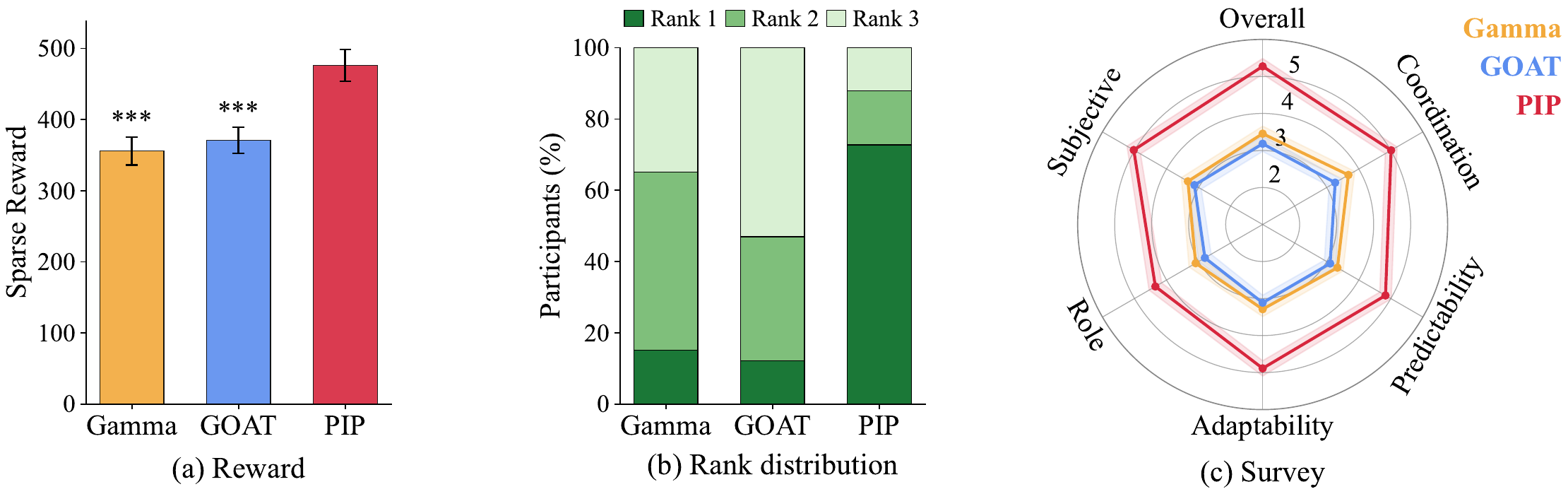}
    \vspace{-5pt}
    \caption{Human evaluation in the Burrito-PO environment. Participants interacted with GAMMA, GOAT, and our method. Results report (a) sparse reward, (b) cooperation preference ranking, and (c) post-interaction survey scores on coordination quality.}
    \vspace{-10pt}
    \label{fig:human_eval}
\end{figure}



As shown in~\cref{fig:human_eval} (a) and (b), participants attain the highest sparse reward with PIP and rank PIP first most often among the three agents. These results indicate that PIP improves task performance while also being preferred as a coordination partner. 
\Cref{fig:human_eval} (c) visualizes the survey results across the six categories. PIP receives the highest score in every category, suggesting that participants perceived PIP as more effective, predictable, adaptable, and desirable to cooperate with under partial observability. Additional analyses are provided in the \textit{appendices}.

\section{Additional Analysis}
\label{sec:analysis}



\subsection{Component Ablation}
\label{sec:ablation}


We conduct an ablation study to isolate the contribution of each component in our framework, as summarized in~\cref{tab:ablation_main}.
Row~(a) corresponds to the GAMMA baseline, whose VAE is trained only on partial-observation trajectories.
Comparing rows~(a) and~(b), replacing this VAE with our student VAE distilled from the joint-view teacher VAE improves performance, suggesting that distillation mitigates representation degradation under partial observability and provides more useful latent partner representations for generated-partner training.
A similar improvement is observed from rows~(e) to~(f), further supporting the benefit of the distilled VAE when Partner-state Belief is used.

Rows~(c) and~(d) add Behavioral Belief and Location Belief, respectively, on top of the distilled VAE, and both improve over row~(b).
This shows that each belief component contributes to ZSC by estimating partner-related information that may be missing from the coordinator's local observation.
Moreover, row~(e) improves over row~(a) even without the distilled VAE, indicating that Partner-state Belief is not only effective under our distilled-VAE training setup but also provides an independent benefit.
Finally, row~(f) achieves the best performance by combining the distilled VAE and Partner-state Belief.
These results show that the components play complementary roles. The distilled VAE improves generated-partner training, while the two belief components provide distinct partner-state estimates, leading to the strongest zero-shot coordination performance when used together. Additional ablations controlling for the increased parameter count and examining the effect of the supervision loss are provided in the \textit{appendices}.

\subsection{Joint-view VAE Representation Analysis}
\label{sec:joint_view_analysis}



\begin{figure}[t]
    \begin{minipage}[t]{0.42\linewidth}
        \captionsetup{type=figure,hypcap=false}
        \centering
        \includegraphics[width=\linewidth]{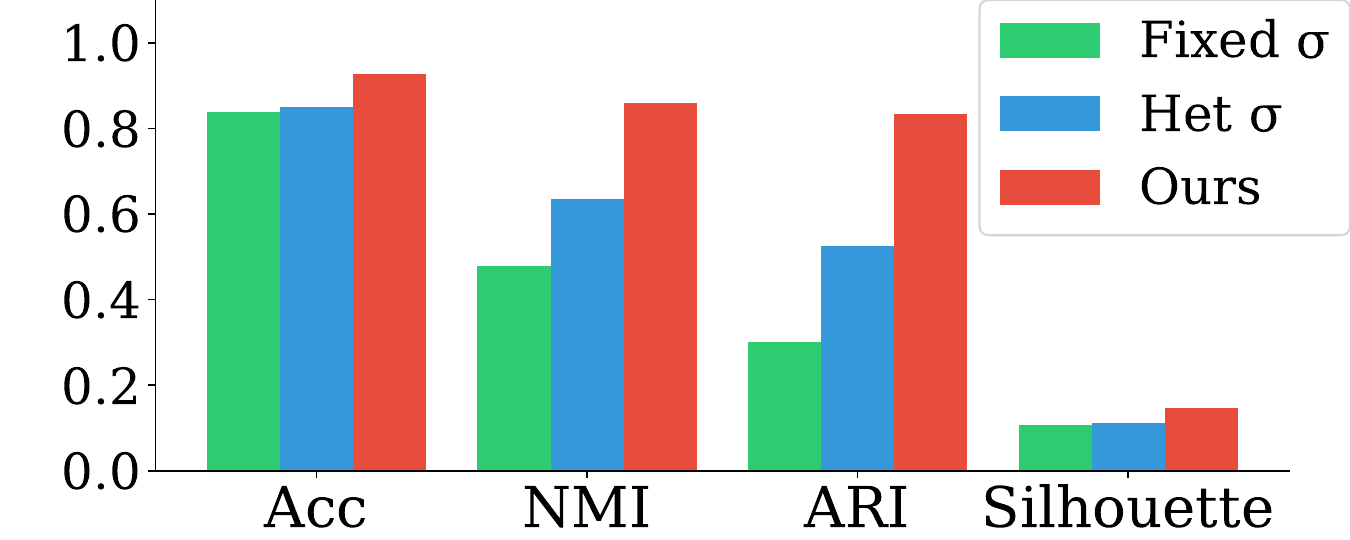}
        \captionof{figure}{Policy classification results.}
        \label{fig:kdvae_analysis1}
    \end{minipage}
    \hfill
    \begin{minipage}[t]{0.54\linewidth}
        \captionsetup{type=figure,hypcap=false}
        \centering
        \includegraphics[width=\linewidth]{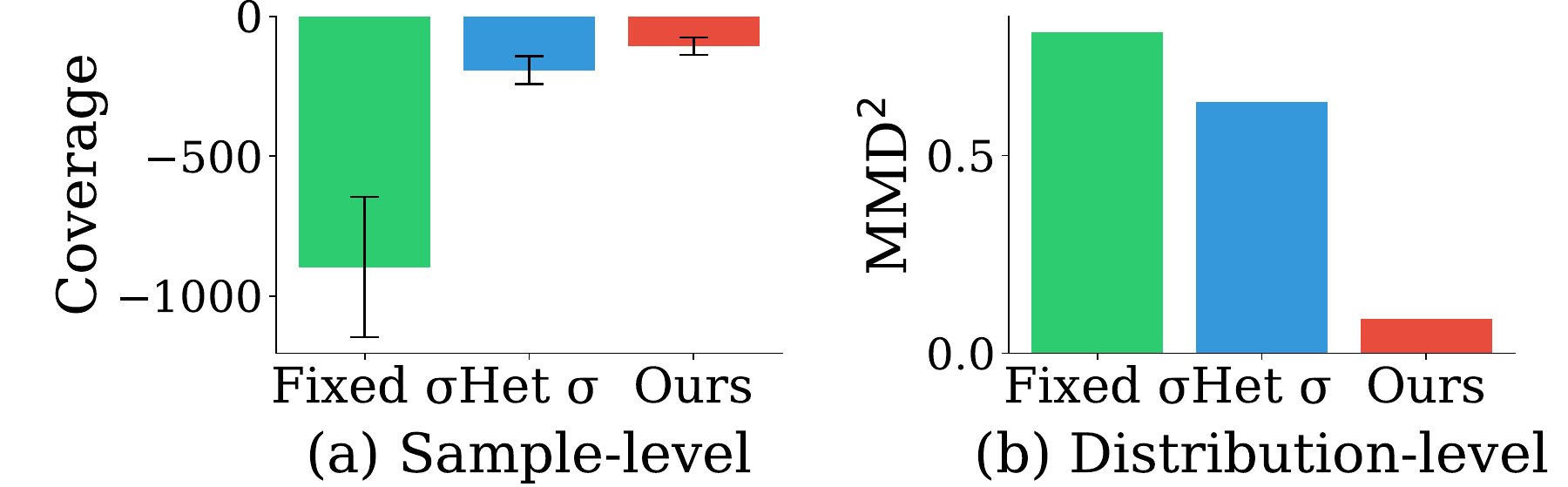}
        \captionof{figure}{Aggregate posterior results.}
        \label{fig:kdvae_analysis2}
    \end{minipage}
     \vspace{-10pt}
\end{figure}

During ZSC inference, the student receives only partner-side inputs, while the teacher is unavailable. Therefore, the student's latent $z$ must satisfy two requirements simultaneously. It must be informative enough to preserve partner identity, and it must remain in-distribution under the prior $p(z)$ from which test-time samples are drawn. The~\cref{fig:kdvae_analysis1} evaluates the first requirement using four standard clustering metrics computed on encoder latents, namely Acc~\citep{grover2018learning}, NMI~\citep{strehl2002cluster, caron2020unsupervised, dang2021nearest}, ARI~\citep{hubert1985comparing, li2021contrastive}, and the silhouette coefficient~\citep{rousseeuw1987silhouettes, yang2017towards}. The distilled student VAE consistently outperforms both the fixed-$\sigma$ and heteroscedastic baselines, with especially large gains on NMI and ARI. The~\cref{fig:kdvae_analysis2} evaluates coverage~\citep{makhzani2015adversarial, tomczak2018vae} using $\log q_{\mathrm{agg}}(z)$ on prior samples $z \sim p(z)$ and the squared MMD~\citep{gretton2012kernel, zhao2019infovae} between $q_{\mathrm{agg}}$ and $p(z)$. PIP achieves both the highest coverage and the lowest MMD$^{2}$. These results show that joint-view distillation produces a partner-discriminative latent representation while reducing prior holes that would place test-time samples outside the decoder's support. Additional metric details and t-SNE visualizations are provided in the \textit{appendices}.

\subsection{Partner-state Belief under Occlusion}
\label{sec:belief_under_occlusion}

\begin{figure}[t]
    \centering
    \begin{minipage}[t]{0.49\linewidth}
        \captionsetup{type=figure,hypcap=false}
        \centering
        \includegraphics[width=\linewidth]{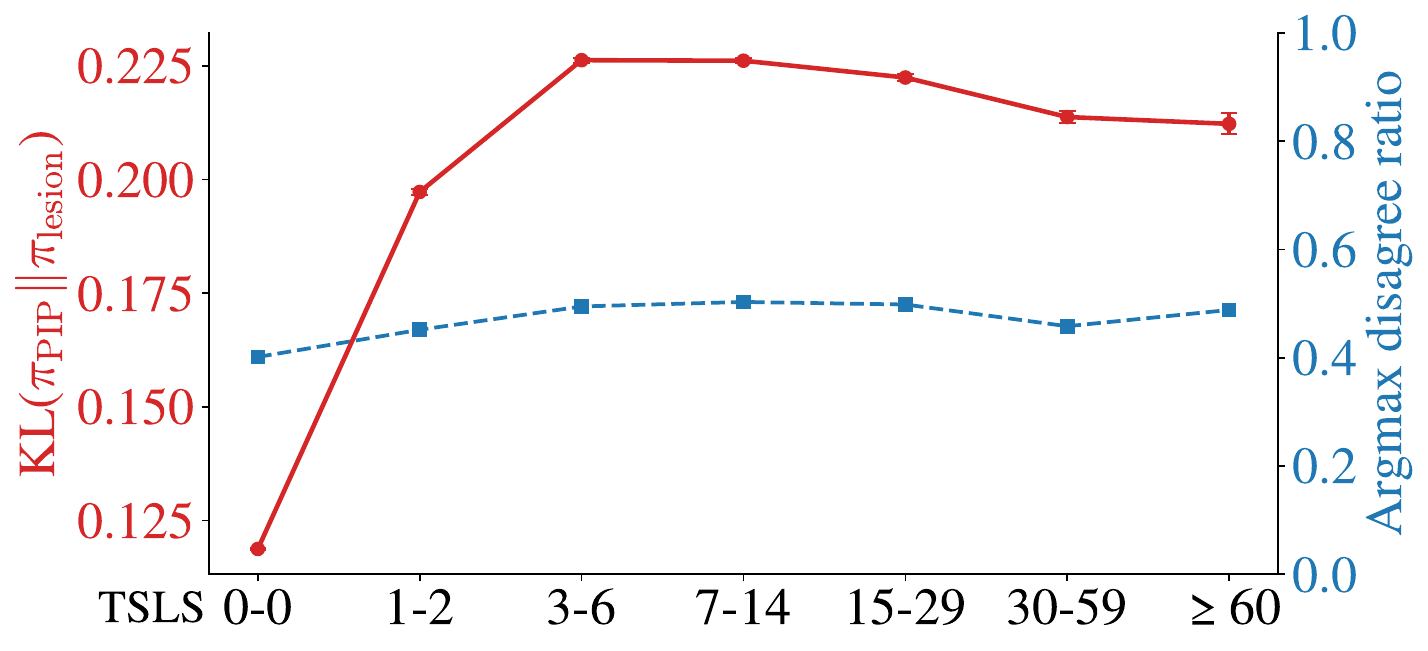}
        \captionof{figure}{Action-distribution difference between $\pi_{\text{PIP}}$ and $\pi_{\text{lesion}}$ across TSLS buckets.}
        \label{fig:belief_effect}
        \vspace{-6pt}
    \end{minipage}
    \hfill
    \begin{minipage}[t]{0.49\linewidth}
        \captionsetup{type=figure,hypcap=false}
        \centering
        \includegraphics[width=\linewidth]{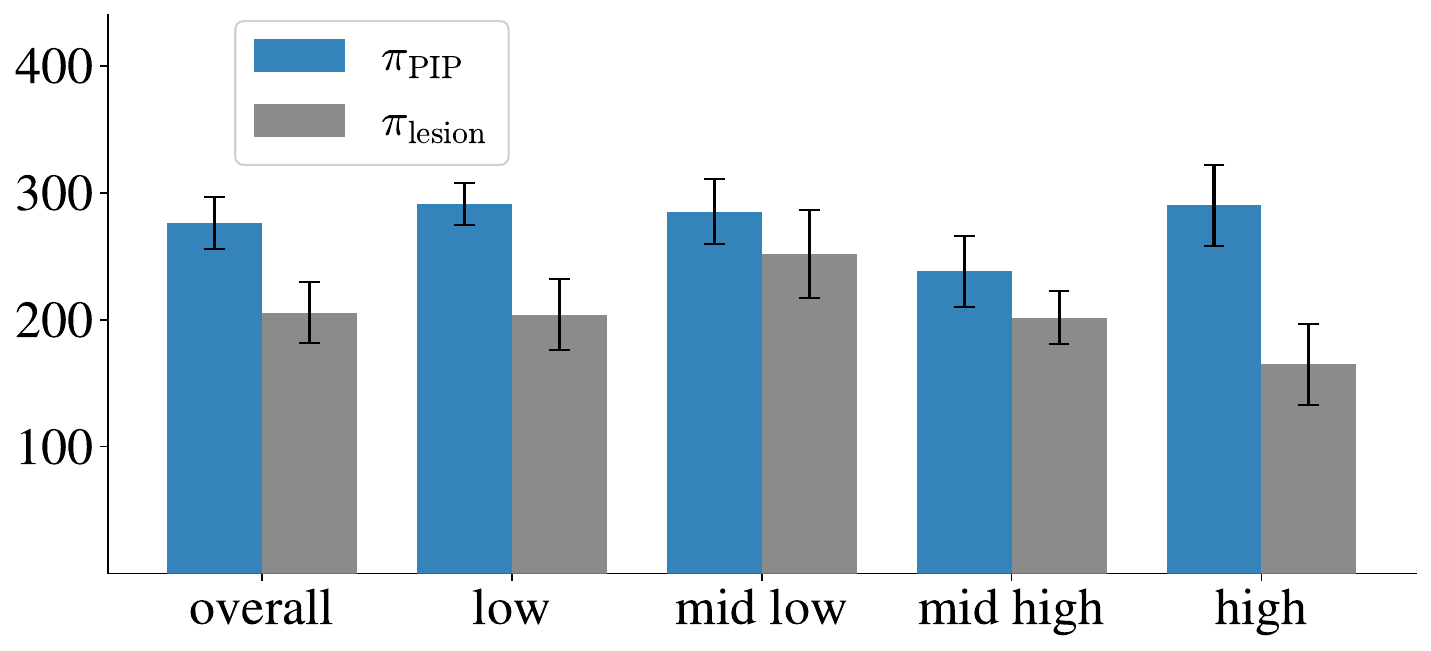}
        \captionof{figure}{Episode sparse reward for PIP and its belief-lesion policy across partner occlusion.}
        \label{fig:belief_zsc_main}
        \vspace{-6pt}
    \end{minipage}
\end{figure}

\paragraph{Effect of Partner-state Belief on Policy Actions.}
We test whether the belief message $\Delta h_t^{\mathrm{act}}$ changes policy actions by comparing $\pi_{\text{PIP}}$ with a lesion policy that zeros the message while preserving the observation and recurrent state. \Cref{fig:belief_effect} shows clear differences in their action distributions across various Time Since Last Seen (TSLS) buckets, the number of steps since the partner left the visible range. The difference is small at TSLS 0, increases after the partner leaves the field of view, and decreases at large TSLS as the hidden state becomes less informative. This suggests that Partner-state Belief most strongly affects actions when inferred partner information remains useful.

\paragraph{Coordination Benefit under Occlusion.}
We analyze how Partner-state Belief benefits coordination across different partner occlusion levels.
We sort the held-out BP policies by the fraction of time their partner is outside the ego agent's field of view when paired with PIP and divide them into four equally sized groups. As shown in~\cref{fig:belief_zsc_main}, PIP consistently outperforms the lesion policy that zero-pads the belief message. The gap is largest for the group whose partner is occluded most frequently. This result shows that Partner-state Belief improves zero-shot coordination most when partner information cannot be obtained directly from the current observation. 
Additional analyses of Partner-state Belief are provided in the \textit{appendices}.

%% file: sec/6_conclusion.tex
\section{Conclusion}
\label{sec:conclusion}
We study generative ZSC under local-view partial observability and propose PIP, which combines a Joint-view VAE for deployment-compatible partner representations with Partner-state Belief networks for hidden-state inference. PIP achieves the highest mean performance across Burrito-PO, Overcooked-PO, and a Melting Pot substrate, with human evaluation and diagnostic analyses further supporting its design. These results demonstrate the importance of partner representations and belief-aware decision making for ZSC under spatially restricted observations.

%% file: sec/X_supple.tex
\appendix

\counterwithin{figure}{section}
\counterwithin{table}{section}
\counterwithin{equation}{section}

\section{Limitations}
Although the deployed policy uses only local observations, the belief module is trained with auxiliary supervision for future partner actions and partner positions. Such labels are available in simulated training environments but may be difficult or costly to obtain in real-world decentralized settings. Future work should therefore investigate weakly supervised or self-supervised belief learning methods that reduce reliance on explicit partner-state labels.

\section{Evaluation Protocol}
\label{sec:appendix:eval_protocol}

\subsection{Construction of BP Agent Candidates}
According to the ZSC-Eval~\citep{wang2024zsc} framework, a well-designed population of behavior preference (BP) agents, constructed with event-based reward functions, can provide behaviorally diverse partners for zero-shot coordination evaluation. Following this idea, we construct, for each domain, a candidate pool of \emph{held-out} BP agents whose sole purpose is to supply diverse evaluation partners that are never seen during target policy training.

\paragraph{Burrito-PO.}
We follow the partner construction protocol used in TALENTS~\citep{li2025talent} and define nine tunable behavior preferences, grouped into three categories of three binary preference dimensions each. Each dimension assigns either an encouraged (a positive weight) or a discouraged (a negative weight) preference to one behavior, and all remaining shaping terms are held fixed and shared across the pool. Enumerating all binary combinations within one category yields $2^3 = 8$ agents, and the three categories together give $24$ BP candidate agents per layout. The preference categories, shaping terms, and corresponding weights are summarized in \cref{tab:bp_features}.

\paragraph{Overcooked-PO.} Overcooked is the domain for which ZSC-Eval originally defines its event set, so we adopt that specification directly rather than designing our own. A BP agent is parameterized by a weight vector over the canonical shaped-event counters, together with one additional entry scaling the sparse task reward. A small number of entries are allowed to take encouraging or discouraging values while the rest are held fixed, and we admit only vectors expressing at most three non-zero preferences, so that each agent exhibits a few interpretable tendencies rather than an arbitrary mixture. From the resulting set of admissible weight vectors we draw $60$ to form the BP candidate pool for this domain. The preference dimensions and their value sets are summarized in~\cref{tab:overcooked_events}.

\paragraph{BP Training phase.}
In both domains, to train the BP agents themselves, the corresponding preference terms are added as auxiliary shaping rewards on top of the sparse environmental reward. Each BP agent is trained together with its corresponding best-response agent, producing partner policies with distinct behavioral tendencies. Once trained, all BP agents are frozen and are used exclusively for selecting evaluation partners and evaluating the target policy. No BP agent, checkpoint, or reward weight is exposed to any stage of target policy training.

\begin{table}[t]
\centering
\begin{tabular}{lc}
\toprule
\textbf{Preference} & \textbf{Reward Shape} \\
\midrule
Plating Ingredients & $-30,\,20$ \\
Washing Plates & $-30,\,20$ \\
Delivering Dishes & $-30,\,20$ \\
\midrule
Chopping Ingredients & $-30,\,20$ \\
Potting Rice & $-30,\,20$ \\
Grilling Meat/Mushroom & $-30,\,20$ \\
\midrule
Taking Mushroom From Dispenser & $-15,\,10$ \\
Taking Rice From Dispenser & $-15,\,10$ \\
Taking Meat From Dispenser & $-15,\,10$ \\
\bottomrule
\end{tabular}
\caption{Event-based BP features and corresponding reward design in Burrito-PO.}
\label{tab:bp_features}
\end{table}

\begin{table}[t]
\centering
\small
\begin{tabular}{lc}
\toprule
\textbf{Events} & \textbf{Weights} \\
\midrule
Put an onion / dish / soup / tomato & \multirow{2}{*}{0} \\
onto the counter &  \\
Pickup an onion / dish / soup / tomato & \multirow{2}{*}{0} \\
from the counter &  \\
Pickup an onion from the onion dispenser & $-20, 0, 10$ \\
Pickup an tomato from the tomato dispenser & $-20, 0, 10$ \\
Pickup a dish from the dish dispenser & $-20, 0, 10$ \\
Pickup a soup & $-20, 0, 5, 10$ \\
Place an ingredient into the pot & $-20, 0, 3, 10$ \\
Deliver a soup & $-20, 0$ \\
Stay & $-0.1, 0, 0.1$ \\
Movement & 0 \\
Order Reward & $0.1, 1$ \\
\bottomrule
\end{tabular}
\caption{Event-based BP features and corresponding reward design in Overcooked-PO}
\label{tab:overcooked_events}
\end{table}

\subsection{Selection of Evaluation BP Agents}

From the BP candidate agents, we select 12 evaluation partners for each layout following the BR-diversity-based selection procedure of ZSC-Eval~\citep{wang2024zsc}.
The goal of this selection step is to obtain a compact evaluation set that preserves behavioral diversity among the candidate partners.

For each candidate BP agent, let $\pi_i^{\mathrm{BP}}$ denote the candidate partner and $\pi_i^{\mathrm{BR}}$ denote its corresponding best-response agent, where $i \in \{1,\dots,N\}$. Here, $N=24$ for Burrito-PO and $N=60$ for Overcooked-PO. We roll out each pair and summarize the behavior of $\pi_i^{\mathrm{BR}}$ using event-based statistics. Let $\mathcal{E}$ be the set of environment events, and let $c_{i,e}^{(r)}$ denote the number of times event $e \in \mathcal{E}$ occurs for the corresponding best-response agent in rollout $r$. The behavior feature of candidate $i$ is defined as the average event-count vector
\[
\phi_i(e) = \frac{1}{R}\sum_{r=1}^{R} c_{i,e}^{(r)},
\qquad e \in \mathcal{E},
\]
where $R$ is the number of rollouts used for estimating the event statistics and set to 5.

We then normalize each event dimension across the candidate pool:
\[
\tilde{\phi}_i(e) =
\frac{\phi_i(e)}
{\max_{j \in \{1,\dots,N\}} \phi_j(e) + \epsilon},
\]
where $\epsilon$ is a small constant for numerical stability. Stacking the normalized feature vectors gives the matrix
\[
\tilde{\Phi} =
\begin{bmatrix}
\tilde{\phi}_1^\top \\
\tilde{\phi}_2^\top \\
\vdots \\
\tilde{\phi}_N^\top
\end{bmatrix}.
\]
We compute the behavioral similarity matrix as
\[
K = \tilde{\Phi}\tilde{\Phi}^{\top}.
\]

Finally, we select a subset $S^\ast$ of $M=12$ evaluation partners by maximizing the log-determinant of the corresponding principal submatrix:
\[
S^\ast =
\arg\max_{S \subset \{1,\dots,N\},\, |S|=M}
\log \det K_S,
\]
where $K_S$ denotes the principal submatrix of $K$ indexed by $S$. Since a larger determinant indicates that the selected behavior feature vectors span a more diverse set of directions, this criterion favors evaluation partners associated with diverse best-response behaviors. The BP agents corresponding to the selected indices $S^\ast$ are fixed as the final held-out evaluation partners for the layout.

\subsection{Zero-shot Evaluation Protocol}

\paragraph{Burrito-PO and Overcooked-PO.} We evaluate the zero-shot coordination performance of the trained coordinator by pairing it with the selected held-out BP partners. The evaluation partner set consists of $12$ BP agents selected from $24$ candidates for Burrito-PO and $60$ candidates for Overcooked-PO for each layout. For each selected BP partner, we run the coordinator-partner pair under both role assignments by swapping their spawn positions. Thus, each partner induces two ordered evaluation pairings.

For each ordered pairing, we collect five episodes, resulting in $12 \times 2 \times 5 = 120$ episodes per evaluation. Equivalently, the coordinator is evaluated for 10 episodes with each selected BP partner, matching the evaluation protocol described in the main paper. All main results and analyses are reported using this selected evaluation set.

We include role-swapped evaluation because partial observability and ego-centric inputs make the two agent roles potentially asymmetric. Although the sparse reward is shared between agents, the policy input representation and dense shaping signals are agent-specific during training. Evaluating the coordinator under both role assignments therefore provides a more complete estimate of its zero-shot coordination performance. 


\paragraph{Melting Pot.} Since Melting Pot provides official bot partners rather than BP agents, we follow a separate protocol for this benchmark. The target policy is paired with each of the three official bots (cooperator, defector, and mixed) on the original $27 \times 27$ \texttt{coop\_mining}, and we collect $100$ episodes per bot, resulting in 300 evaluation episodes in total. In each episode, we measure the episodic reward obtained by the target policy, and we report its mean and standard error.

\section{Implementation Details}
\label{sec:appendix:impl_detail}

\subsection{Window Size Selection}
\label{subsec:vs}

\begin{figure}[t]
\centering
\includegraphics[width=.70\linewidth]{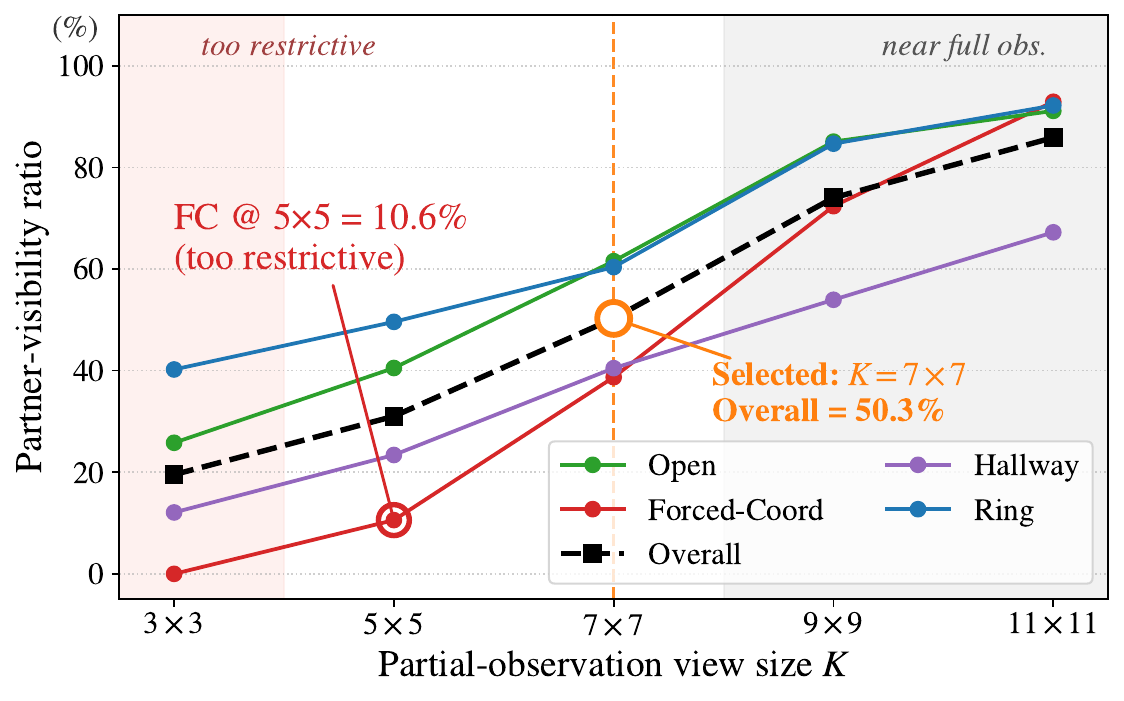}
\caption{Partner-visibility ratio (\%) by view size and layout, measured from Self-Play rollouts using the MEP algorithm.}
\label{fig:vs_layout}
\end{figure}

Zero-shot coordination (ZSC) inherently requires agents to collaborate with a partner. When the partner is almost never observable, the task effectively degenerates into single-agent play, eliminating meaningful coordination signals.
Conversely, when the partner is almost always visible, the setting becomes nearly indistinguishable from full observability. This makes it difficult to evaluate zero-shot coordination performance under partial observability.
To determine an appropriate PO window size for Burrito-PO, we first empirically measured the partner-visibility ratio for different view sizes across all Burrito-PO layouts. Specifically, we performed self-play rollouts using SP agents trained in Stage 1 of MEP, running 5 episodes per policy (40 episodes per layout in total), and recorded partner visibility at each timestep for each view size. The results are summarized in~\cref{fig:vs_layout}. For view size $K = 3 \times 3$, the visibility ratio in the Forced-Coord (FC) layout is $0\%$, making it overly restrictive as a coordination task. While $K = 5 \times 5$ achieves moderate visibility in other layouts, it remains insufficient in FC to reliably evaluate PO-ZSC performance.
In contrast, for $K = 9 \times 9$ and $K = 11 \times 11$, visibility ratios are consistently high across all layouts. As a result, the distinction between full and partial observability becomes blurred, making it difficult to isolate zero-shot coordination performance under partial observability.
Based on this analysis, we select $K = 7 \times 7$ as the default window size for Burrito-PO, as it maintains a non-trivial level of visibility across all layouts while preserving sufficient occlusion. Under this setting, agents must meaningfully infer their partners while still receiving intermittent observations, forming an appropriate regime for evaluating ZSC under partial observability.
We follow a similar procedure for the remaining benchmarks. This yields $K = 5 \times 5$ for Overcooked-PO. For Melting Pot, we adopt the official egocentric view range (5 cells to each side, 1 cell behind, and 9 cells ahead of the agent, with each cell rendered at $8 \times 8$ pixels) which we found to satisfy the criterion above, exhibiting neither degenerate near-zero nor near-full partner visibility across substrates. We therefore keep it unchanged and use these settings throughout our experiments.

\subsection{Hyperparameter Selection}
Unless otherwise specified, hyperparameters are adopted from the corresponding baseline methods whenever possible, and the remaining hyperparameters are selected via grid search over a predefined search space, taking computational budget into consideration.

\subsection{Reward Function}
In the Burrito environment, the one-step reward consists of a shared sparse reward and an agent-specific dense shaping reward. After each joint action, the reward for agent $i$ is computed as
\begin{equation}
r_i^t = r_{\mathrm{sparse}}^t + \eta_t r_{\mathrm{dense}, i}^t ,
\label{eq:burrito_step_reward}
\end{equation}
where $r_{\mathrm{sparse}}^t$ is the cooperative sparse reward shared by both agents, $r_{\mathrm{dense}, i}^t$ is the dense shaping reward caused by agent $i$ at timestep $t$, and $\eta_t$ is the reward-shaping coefficient.

\paragraph{Sparse reward.}
The sparse reward is generated only when a dish is delivered. If no delivery occurs, the sparse reward is zero. When a delivery occurs, the delivery reward is divided equally between the two agents, so both agents receive the same sparse reward regardless of which agent performed the final delivery action:
\begin{equation}
r_{\mathrm{sparse}}^t =
\begin{cases}
\frac{1}{2} R_{\mathrm{del}}^t, & \text{if a dish is delivered at timestep } t, \\
0, & \text{otherwise}.
\end{cases}
\label{eq:burrito_sparse_reward}
\end{equation}

The delivery reward depends on the delivered dish type. If the dish matches the first active order, it is treated as an in-order delivery. If it matches another order in the order list, it is treated as an out-of-order delivery. Otherwise, it is treated as a random dish delivery. The delivery reward is computed as
\begin{equation}
R_{\mathrm{del}}^t =
\begin{cases}
20 + 20 n_{\mathrm{ing}} + m c_{\mathrm{tip}}, & \text{in-order delivery}, \\
20 + 20 n_{\mathrm{ing}} + m c_{\mathrm{tip}}, & \text{out-of-order delivery}, \\
R_{\mathrm{random}}, & \text{random dish delivery},
\end{cases}
\label{eq:burrito_delivery_reward}
\end{equation}
where $n_{\mathrm{ing}}$ is the number of ingredients, $m$ is the consecutive-order multiplier, $c_{\mathrm{tip}}$ is a deadline-based tip, and $R_{\mathrm{random}}$ is the fixed reward for random dish delivery. In-order deliveries increase the consecutive-order count, whereas out-of-order deliveries reset it.

\paragraph{Dense shaping reward.}
The dense reward provides immediate feedback for useful intermediate task progress, such as cooking ingredients, assembling food, serving dishes, washing dishes, and resolving hazards. Unlike the sparse reward, the dense reward is not shared. Each agent receives dense reward only for shaping events caused by its own action at the current timestep.

The dense shaping term is used as a training aid. The coefficient $\eta_t$ is annealed toward zero during training, so dense rewards help exploration and credit assignment early in training, while the final policy is optimized primarily toward the sparse cooperative delivery objective.

\subsection{VAE Dataset Construction}

\subsubsection{Burrito-PO.}

Following the VAE data-collection protocol of Burrito~\citep{li2025talent}, we adapt the dataset construction to our training population and collect trajectories at the level of ordered policy pairs. For VAE training, we use 24 MEP policies as the training population and enumerate all ordered pairs by taking the Cartesian product of this population with itself. This produces $24 \times 24 = 576$ ordered policy pairs, including self-pairs. We collect three episodes for each ordered pair, resulting in 1728 trajectories, which are used to train the generative partner model.

\subsubsection{Overcooked-PO.}

Following the VAE data-collection protocol of GAMMA~\citep{liang2024gamma}, we adapt the dataset construction to our setting and collect trajectories using our training population. For VAE training, we use 24 MEP policies as the training population and collect 20k trajectories, which are used to train the generative partner model.

\subsubsection{Melting Pot.}

Because the original Melting Pot work~\citep{leibo2021scalable} does not provide a separate protocol for constructing a VAE training dataset, we follow the Burrito-PO protocol and adapt it to the Melting Pot setting. For VAE training, we use 24 MEP policies as the training population and enumerate all ordered pairs by taking the Cartesian product of this population with itself. This produces $24 \times 24 = 576$ ordered policy pairs, including self-pairs. We collect four episodes for each ordered pair, resulting in 2304 trajectories, which are used to train the generative partner model.

\subsection{Baselines}

\noindent\textbf{FCP}~\citep{strouse2021fcp}
builds a frozen partner pool by collecting init / mid / final checkpoints from several self-play seeds, and trains a best-response (BR) ego policy by uniformly sampling partners from this pool. Our only strict-PO modification is the observation regime. The original method was evaluated under full observability, whereas in our setting both the Stage~1 self-play training (8 seeds) and the Stage~2 BR training are run with ego-centric partial observations using view radius~3, corresponding to a $7 \times 7$ local observation window. Thus, neither the ego nor any frozen partner observes the other agent's state unless it falls within the local field of view. The algorithm itself (skill-stratified pool construction, uniform partner sampling, PPO-based BR training) is left unchanged.

\noindent\textbf{MEP}~\citep{zhao2023maximum} builds a frozen partner pool by collecting init / mid / final checkpoints from several self-play seeds with a population entropy bonus, and trains a best-response (BR) ego policy using prioritized sampling that biases sampling toward low-performing partners. Our only strict-PO modification is the observation regime. The original method was evaluated under full observability, whereas in our setting both the Stage~1 self-play training (8 seeds) and the Stage~2 BR training are run with ego-centric partial observations. Thus, neither the ego nor any frozen partner observes the other agent's state unless it falls within the local field of view. The algorithm itself (self-play with a population entropy bonus, prioritized partner sampling, PPO-based BR training) is left unchanged.

\noindent\textbf{E3T}~\citep{yan2023e3t} avoids any external partner pool by setting the partner side to a mixture of an ego self-copy and a random policy, $\pi_p = \epsilon\,\pi_r + (1 - \epsilon)\,\pi_e$, and jointly trains, end-to-end, a context encoder consuming a length-$k$ history of (state, partner-action) pairs together with a head that predicts the partner's next action distribution; the ego policy is conditioned on the current state and the predicted partner action distribution and trained by PPO. Our strict-PO setting introduces two modifications. First, every state input on both the encoder side and the ego-policy side is replaced with the ego's partial observation, so that no module ever sees the environment state directly. Second, the partner-action sequence fed to the encoder is FOV-masked: steps where the partner was inside the ego's field of view keep the true action token, while steps where the partner was outside are replaced by a learnable $\langle\textsc{unk}\rangle$ embedding, since the original lossless-encoding assumption no longer holds when the partner leaves the ego's view. All other components (mixture partner, encoder/ego separate optimization, single-stage end-to-end) are kept as in the original.

\noindent\textbf{ERS}~\citep{muglich2025expected}
learns expected-return symmetries, policy transformations $\sigma$ composed of an observation-side input map and an action-side output map that approximately preserve the expected return, and trains a cooperator against symmetry-transformed partners. In our setting, we train $\sigma_{ij}$ for all 56 ordered pairs of eight strict-PO self-play base policies by pairing the frozen $\text{base}_i$ with $\sigma \circ \text{base}_j$ and maximizing the cross-play return with PPO, where only $\sigma$ and a fresh critic are trainable. We retain the top $l=8$ symmetries ranked by cross-play return, compose a 16-policy pool from the base policies and their symmetry-transformed counterparts, and train the cooperator with PPO against uniformly sampled pool partners. We reduce the population size and pair budget relative to the original setting for computational reasons while keeping the symmetry architecture and training procedure unchanged.

\begin{table}[t]
\centering
\begin{tabular}{lccc}
\toprule
\textbf{Hyperparameter} & \textbf{Burrito-PO} & \textbf{Overcooked-PO} & \textbf{Melting Pot} \\
\midrule
CNN kernels                 & [3, 3], [3, 3], [3, 3] & [3, 3], [3, 3], [3, 3] & [8, 8], [4, 4], [3, 3] \\
CNN channels                & \multicolumn{3}{c}{[32, 64, 32]} \\
hidden layer size           & \multicolumn{3}{c}{[64]} \\
recurrent layer size        & \multicolumn{3}{c}{64} \\
activation function         & \multicolumn{3}{c}{ReLU} \\
weight decay                & \multicolumn{3}{c}{0} \\
\midrule
environment steps           & 60M  & 100M & 200M \\
parallel environments       & 64   & 200  & 128 \\
episode length              & 2400 & 400  & 1000 \\
PPO batch size              & $2 \times 2400 \times 64$ & $2 \times 400 \times 200$ & $2 \times 1000 \times 128$ \\
mini batch size             & $2 \times 2400 \times 64 \div 10$ & $2 \times 400 \times 200$ & $2 \times 1000 \times 128 \div 16$ \\
\midrule
PPO epoch                   & \multicolumn{3}{c}{15} \\
PPO learning rate           & \multicolumn{3}{c}{0.0005} \\
GAE $\lambda$               & \multicolumn{3}{c}{0.95} \\
discounting factor $\gamma$ & \multicolumn{3}{c}{0.99} \\
\bottomrule
\end{tabular}
\caption{Hyperparameters for policy models in each environment.}
\label{tab:policy_hparams_all}
\end{table}

\begin{table}[t]
\centering
\begin{tabular}{lc}
\toprule
\textbf{Hyperparameter} & \textbf{Value} \\
\midrule
\multicolumn{2}{l}{\textit{\textbf{Behavioral Belief}}} \\
$\lambda_{\mathrm{behavior}}$ & 0.1 \\
$K_{\mathrm{off}}$ & 10 \\
$\gamma_{\mathrm{rec}}$ & 0.90 \\
$\gamma_{\mathrm{ten}}$ & 0.97 \\
$\alpha_{\mathrm{rec}}$ & 1.0 \\
$\alpha_{\mathrm{ten}}$ & 0.3 \\ \midrule
\multicolumn{2}{l}{\textit{\textbf{Location Belief}}} \\
$\lambda_{\mathrm{location}}$ & 1.0 \\
$\alpha$ & 0.4 \\
Gaussian target sigma & 1.5 \\
$H_{\mathrm{max}}$ & 50 \\
$\Delta$ & [0, 1, 3, 5, 10, 20] \\
$w_\Delta$ & [1.0, 0.5, 0.7, 1.0, 1.0, 0.8] \\
$\lambda_{\mathrm{aux}}$ & 0.1 \\
\bottomrule
\end{tabular}
\caption{Partner-state Belief hyperparameters for target policy models.}
\label{tab:target_policy_hparams}
\end{table}

\begin{table}[!t]
    \centering
    \small
    \setlength{\tabcolsep}{3pt}
    \begin{tabular}{lccc}
    \toprule
    Metric & Baseline & PIP (Ours) & Ratio \\
    \midrule
    Total parameters & 463K & 591K & 1.28$\times$ \\

    Actor rollout inference time & 1.13 ms & 2.05 ms & 1.81$\times$ \\
    Actor rollout FLOPs & 0.381G & 0.764G & 2.01$\times$ \\
    \bottomrule
    \end{tabular}
    \caption{Representative computational costs at the training rollout batch size of 64. Baseline corresponds to GAMMA / GOAT, whose target policies share an identical architecture.}
    \label{tab:com}
\end{table}

\noindent\textbf{GAMMA}~\citep{liang2024gamma}
pretrains a conditional VAE that generates partner behaviors, and trains the cooperator with PPO by sampling a fresh latent $z \sim \mathcal{N}(0, I)$ at the start of each episode and rolling out against the decoder conditioned on $z$. Our strict-PO setting introduces two modifications. First, while the original VAE encoder consumes full joint trajectories $\tau$ and the decoder relies on simulator-side state information, our VAE is retrained so that both the encoder and the decoder take only the partner's PO as input; the partner generator is therefore closed under partial observability, and the partner behaviors produced by the decoder remain within the action distribution of strict-PO policies. Second, although the original work places no explicit restriction on the cooperator's critic, we disable the centralized value function, so that both the actor and the critic of the cooperator operate solely on its own PO, i.e.\ in IPPO mode. The remaining structure (per-episode $z$ sampling, BR training of the cooperator against the $z$-conditioned partner via PPO) is identical to the original.

\noindent\textbf{GOAT}~\citep{chaudhary2026goat} reuses the same frozen VAE decoder as a partner generator but replaces the fixed prior $z \sim \mathcal{N}(0, I)$ with an adversary policy $\pi_A$ that maps $z \sim \mathcal{N}(0, I) \to z' = \pi_A(z)$ so as to maximize the regret $\mathrm{Regret}(\pi_P, \pi_C) = \mathbb{E}\!\left[V(\pi_P, \pi_P) - V(\pi_P, \pi_C)\right] = J_{\text{SP}} - J_{\text{XP}}$. Each episode therefore consists of a two-phase rollout, a partner-vs-partner self-play rollout that estimates $J_{\text{SP}}$ and a partner-vs-cooperator cross-play rollout that estimates $J_{\text{XP}}$, with the cooperator and the adversary forming the two sides of a minimax game. Our strict-PO setting inherits the two GAMMA modifications and adds nothing else: the partner-generator VAE is trained to consume only the partner's partial observation, and the cooperator is trained as IPPO. Since the adversary $\pi_A$ operates purely on the latent $z$ and never receives any observation as input, it does not introduce any new PO-violating surface; the two-phase rollout structure and the regret objective are kept exactly as in the original method.

\begin{table*}[t]
\centering
\begin{tabular}{lcccccc}
\toprule
\textbf{hyperparameter} & \multicolumn{3}{c}{\textbf{Stage 1 (teacher pretrain)}} & \multicolumn{3}{c}{\textbf{Stage 2 (student distill)}} \\
\midrule
trained module & \multicolumn{3}{c}{teacher (enc + dec)} & \multicolumn{3}{c}{student (enc + dec), teacher frozen} \\
encoder input & \multicolumn{3}{c}{$(o_1 \cup o_2) \oplus o_\text{partner}$} & \multicolumn{3}{c}{$o_\text{partner}$} \\
$\lambda_{KD}$ & \multicolumn{3}{c}{--} & \multicolumn{3}{c}{10.0} \\
epoch & \multicolumn{3}{c}{500} & \multicolumn{3}{c}{500} \\
\midrule
\multicolumn{7}{c}{shared across stages} \\
\midrule
\textbf{hyperparameter} & \multicolumn{2}{c}{\textbf{Burrito-PO}} & \multicolumn{2}{c}{\textbf{Overcooked-PO}} & \multicolumn{2}{c}{\textbf{Melting Pot}} \\
\midrule
CNN kernels & \multicolumn{2}{c}{[3, 3], [3, 3], [3, 3]} & \multicolumn{2}{c}{[3, 3], [3, 3], [3, 3]} & \multicolumn{2}{c}{[8, 8], [4, 4], [3, 3]} \\
CNN channels & \multicolumn{6}{c}{[32, 64, 32]} \\
hidden layer size & \multicolumn{6}{c}{[256]} \\
recurrent layer size & \multicolumn{6}{c}{256} \\
activation function & \multicolumn{6}{c}{ReLU} \\
weight decay & \multicolumn{6}{c}{0.0001} \\
episode length & \multicolumn{2}{c}{2400} & \multicolumn{2}{c}{400} & \multicolumn{2}{c}{1000} \\
chunk length & \multicolumn{2}{c}{200} & \multicolumn{2}{c}{100} & \multicolumn{2}{c}{100} \\
chunks per episode & \multicolumn{2}{c}{6} & \multicolumn{2}{c}{1} & \multicolumn{2}{c}{5} \\
batch size & \multicolumn{6}{c}{256} \\
learning rate & \multicolumn{6}{c}{0.001} \\
KL penalty coefficient $\beta$ & \multicolumn{6}{c}{$0 \rightarrow 0.1$} \\
latent variable dimension & \multicolumn{6}{c}{16} \\
\bottomrule
\end{tabular}
\caption{Hyperparameters for the Joint-view VAE.}
\label{tab:vae_hparams}
\end{table*}

\subsection{Policy and VAE model details}

We use a convolutional recurrent backbone for all policy models, where observations are encoded by a three-layer CNN, followed by a fully connected layer and a recurrent layer for partial observability. The MEP partners are trained with PPO using this backbone.
The policy backbones and PPO hyperparameters for Burrito-PO, Overcooked-PO, and Melting Pot are summarized 
in~\cref{tab:policy_hparams_all}.

The target policy differs from the MEP models in two aspects. First, it is trained with simulated partner data generated from the pretrained partner model, rather than through the original MEP population training procedure. Second, our target coordinator additionally includes the Partner-state Belief branch described in the main paper. Except for this additional branch, the base policy backbone and PPO optimization hyperparameters are kept identical to those of the MEP partners. The hyperparameters used for the Partner-state Belief branch during target policy training are summarized in~\cref{tab:target_policy_hparams}. The reverse KL direction discourages probability mass in low-feasibility regions without requiring the prediction to reproduce the full spread of the smoothed target.

For the generative partner model, we train the Joint-view VAE in two stages. In the first stage, the teacher VAE is trained using a richer partial-view input constructed from the union of both agents' partial observations and the partner-side observation. In the second stage, the teacher is frozen and the student VAE is trained from the partner-side partial observation with an additional posterior distillation loss. We use $\mathrm{KL}(q_T | q_S)$ rather than the reverse direction so that the student posterior covers all modes of the more informative teacher posterior, preserving the diversity that the teacher captures from the joint view. Both stages share the same backbone architecture, optimizer settings, latent dimension, and KL annealing schedule. The VAE hyperparameters are provided in~\cref{tab:vae_hparams}.

\section{Computational Resources}
\label{sec:appendix:computational}
We report representative computational costs in \cref{tab:com}, measured at the training rollout batch size of 64. Compared with the baseline average of GAMMA and GOAT, our method increases total parameters from 463K to 591K, with the additional parameters concentrated in the actor while the critic size remains unchanged. Actor rollout inference time increases from 1.13 ms to 2.05 ms, and actor rollout computation increases from 0.381G to 0.764G FLOPs. This corresponds to an actor rollout inference rate of roughly 500 Hz at this batch size, suggesting that the inference overhead is unlikely to preclude real-time execution in this setting.

\section{Environment Layouts}

\subsection{Burrito-PO}



\begin{figure}[t]
\centering

\begin{subfigure}[t]{0.47\linewidth}
\centering
\includegraphics[width=\linewidth,height=0.20\textheight,keepaspectratio]{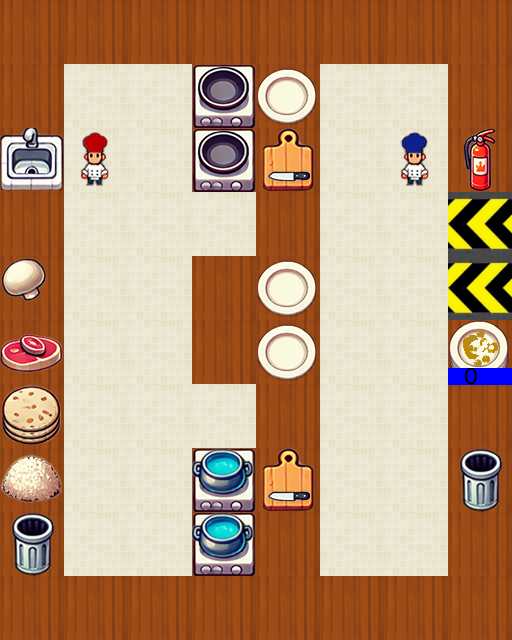}
\caption{Forced-Coord}
\label{fig:fc}
\end{subfigure}
\hfill
\begin{subfigure}[t]{0.47\linewidth}
\centering
\includegraphics[width=\linewidth,height=0.20\textheight,keepaspectratio]{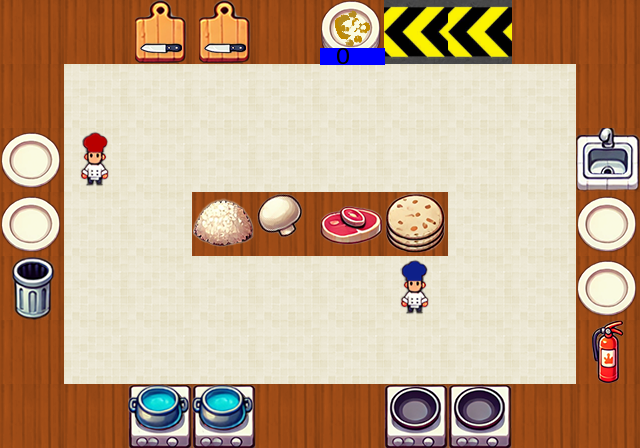}
\caption{Open}
\label{fig:open}
\end{subfigure}

\vspace{0.3em}

\begin{subfigure}[t]{0.47\linewidth}
\centering
\includegraphics[width=\linewidth,height=0.20\textheight,keepaspectratio]{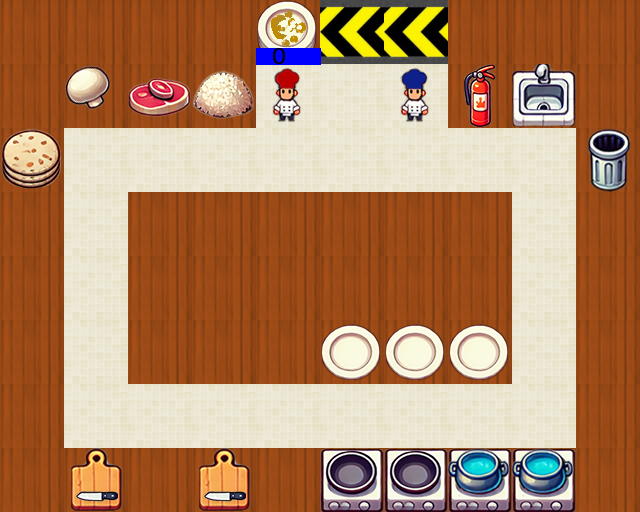}
\caption{Ring}
\label{fig:ring}
\end{subfigure}
\hfill
\begin{subfigure}[t]{0.47\linewidth}
\centering
\includegraphics[width=\linewidth,height=0.20\textheight,keepaspectratio]{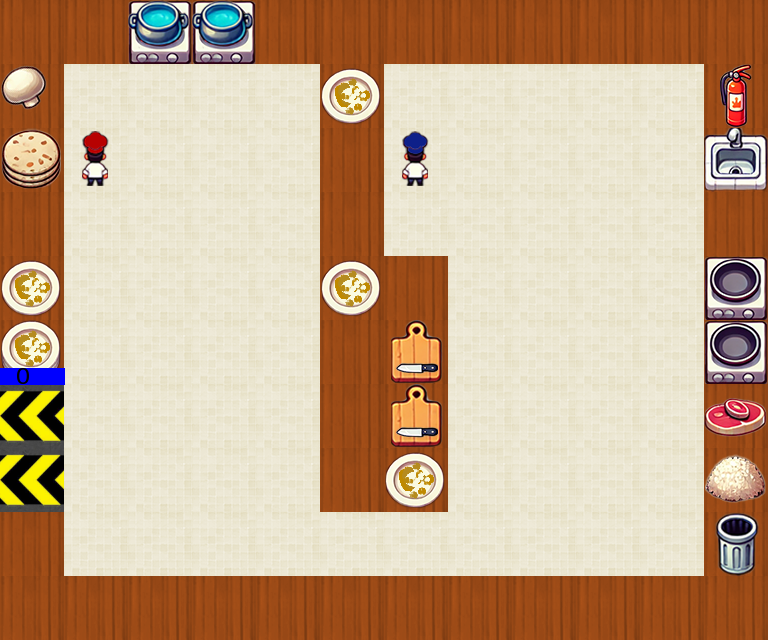}
\caption{Hallway}
\label{fig:burrito}
\end{subfigure}

\caption{Burrito environment layouts used in our experiments. Each layout induces different coordination patterns under ego-centric partial observability.}
\label{fig:layouts}
\end{figure}

We evaluate all methods on four Burrito-PO layouts, shown in~\cref{fig:layouts}, which are adapted from Burrito~\citep{li2025talent}. The layouts differ in workspace openness, movement bottlenecks, object placement, and the spatial separation between task-relevant stations, thereby inducing different partial-observation challenges. The Forced-Coord, Open, Ring, and Hallway layouts correspond to panels (a)--(d) in~\cref{fig:layouts}, respectively.

\noindent\textbf{Forced-Coord (FC)} consists of two narrow vertical workspaces separated by central obstacles and counters. Each agent has access to different nearby task objects, and the usable movement paths are constrained by narrow passages. This layout explicitly enforces spatial separation between agents, making the full recipe infeasible for a single agent under the same task constraints. Therefore, we do not report a single-play shortest-path completion step count for this layout. The layout instead emphasizes role division and handoff-style coordination, while the spatial separation makes it difficult for one agent to continuously observe the other.

\noindent\textbf{Open} provides a large central workspace with relatively few movement constraints. Ingredient dispensers, pots, plates, and delivery-related objects are placed around the boundary of the open area. Since agents can move more freely and visual contact is easier to recover, this layout represents the least structurally constrained setting among the four maps.

\noindent\textbf{Ring} contains a central blocked region surrounded by a loop-shaped traversable corridor. Task objects are distributed around the outside of the loop, so agents must coordinate while moving along constrained circular routes. The loop structure limits possible movement directions and often causes agents to disappear behind the central obstacle under ego-centric partial observation.

\noindent\textbf{Hallway} has a larger and more irregular workspace with task stations distributed across separated regions. The layout includes longer navigation paths and multiple object clusters, requiring agents to coordinate across delayed and partially occluded information. Compared with the other maps, Hallway introduces more complex spatial organization and stronger demands on navigation and partner-state inference.

The single-play shortest-path completion steps further quantify the structural differences among the layouts where a solo completion path is feasible. Open requires 137 steps, Ring requires 156 steps, and Hallway requires 172 steps to complete the recipe under a shortest-path solo planner. This increase reflects that the layouts require progressively longer navigation paths, more distributed object access, and more complex task execution even before multi-agent coordination is considered. Thus, the layouts differ not only in visual shape but also in the amount of movement and planning required to complete the underlying cooking task.


These layout variations allow us to evaluate coordination under different forms of partial observability, ranging from forced spatial separation and relatively open interaction spaces to loop-like movement and spatially distributed task structures.

\subsection{Overcooked-PO}

\begin{figure}[t]
    \centering
    \includegraphics[width=0.60\linewidth]{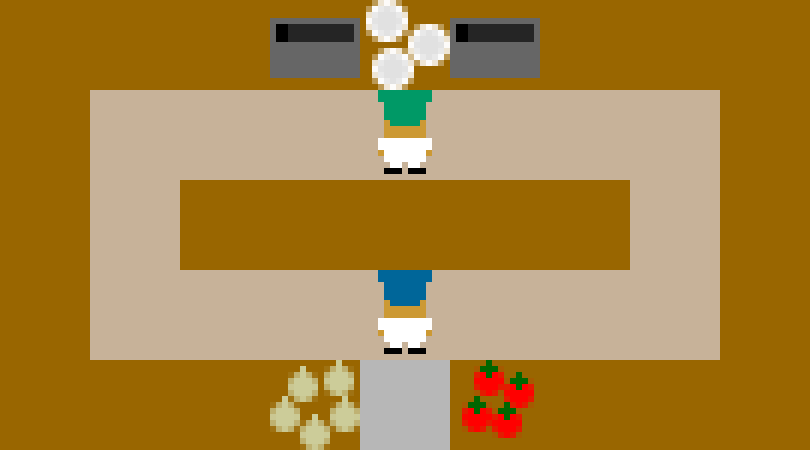}
    \caption{The Overcooked environment layout used in our experiments, counter circuit.}
    \label{fig:overcooked_countercircuit}
\end{figure}

\begin{figure}[t]
    \centering
    \begin{subfigure}[b]{0.35\linewidth}
        \centering
        \includegraphics[width=\linewidth]{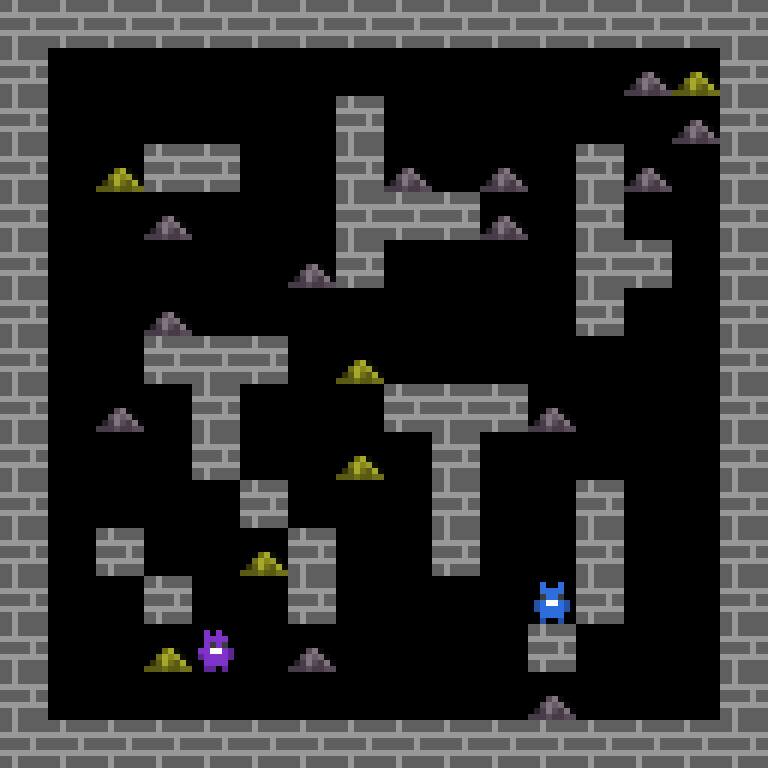}
        \caption{\texttt{coop\_mining\_small} ($16 \times 16$)}
        \label{fig:coop_mining_small}
    \end{subfigure}
    \hspace{0.04\linewidth}
    \begin{subfigure}[b]{0.35\linewidth}
        \centering
        \includegraphics[width=\linewidth]{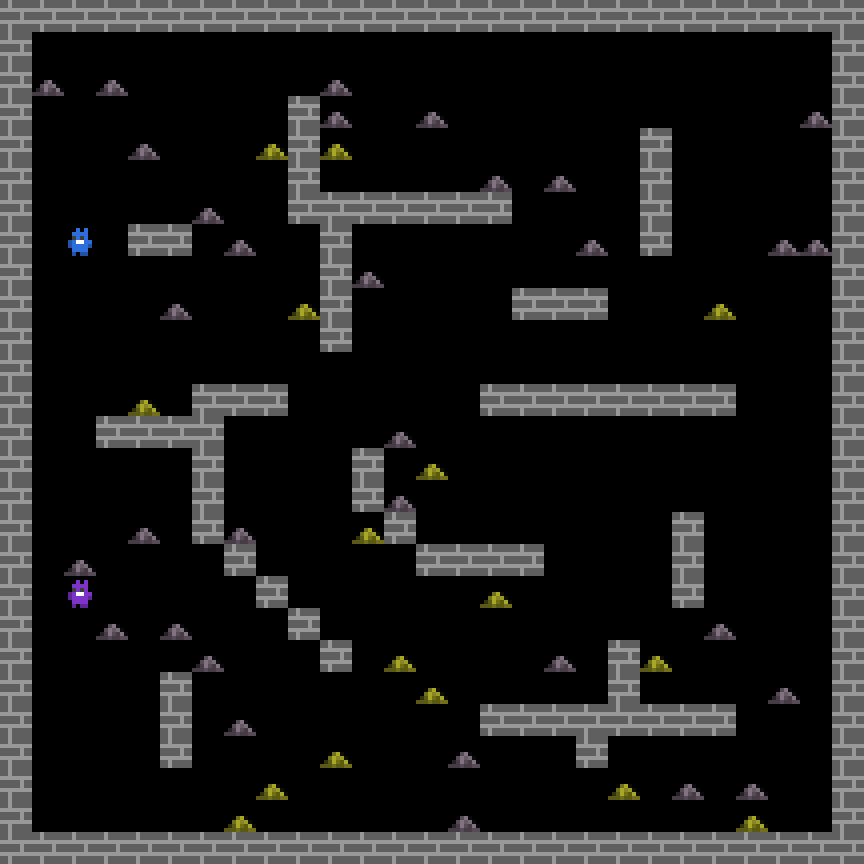}
        \caption{\texttt{coop\_mining} ($27 \times 27$)}
        \label{fig:coop_mining_big}
    \end{subfigure}
    \caption{The Melting Pot coop mining maps used in our experiments. Grey walls occlude the agents' view, iron ore (grey) gives $+1$ when mined alone, whereas gold ore (yellow) gives $+8$ to each agent only if both mine the same ore within a 3-step window.}
    \label{fig:coop_mining}
\end{figure}

We evaluate on the \texttt{counter\_circuit} layout, a $5 \times 9$ ring shaped kitchen whose central counter block leaves only a one tile wide circular corridor. The two cooks cannot pass each other and lose sight of one another whenever they are on opposite sides of the ring. We chose this layout because it contains a tomato dispenser in addition to the onion dispenser. All five layouts of the original ZSC benchmark, \texttt{cramped\_room}, \texttt{asymmetric\_advantages}, \texttt{coordination\_ring}, \texttt{forced\_coordination}, and the onion only \texttt{counter\_circuit\_1\_order}, expose onions only and admit a single fixed recipe. As a result, partner uncertainty is reduced to a purely spatial problem. We instead use the \texttt{counter\_circuit}, which preserves the same ring topology while adding tomatoes. In this layout, all three recipes, onion tomato tomato, onion onion tomato, and onion tomato, require both ingredients. An agent must therefore infer which recipe its partner is pursuing from the pot contents and from which dispenser the partner visits. This makes the layout a benchmark for both ingredient level and spatial partner inference under partial observability.

\subsection{Melting Pot}

In \texttt{coop\_mining}, iron ore can be mined individually for a reward of $+1$, whereas gold ore yields $+8$ to each agent only when both agents mine the same ore within a three step window. Obtaining the high value reward therefore requires successful rendezvous with a partner who is typically outside the agent's egocentric field of view. Our pipeline trains an MEP partner population, a partner generating VAE, a KD teacher student VAE, and finally the target policy. 
Each stage requires hundreds of millions of environment steps. For our method, training on the original $27 \times 27$ map (global map size $216 \times 216$) results in out-of-memory errors on an NVIDIA RTX A6000 GPU, making the computational and memory requirements prohibitively expensive.
We therefore train all stages on a reduced $16 \times 16$ variant, \texttt{coop\_mining\_small}, with a global map size of $128 \times 128$, which preserves the original dynamics and reward rules. Since the policy input remains a fixed $88 \times 88 \times 3$ egocentric RGB observation and the action space is identical across map sizes, the learned policies transfer without modification. We evaluate them in a zero shot manner on the original $27 \times 27$ \texttt{coop\_mining} map against the official Melting Pot bots, cooperator, defector, and mixed.

\subsection{Discussion of Environments}

Our environment selection preserves comparability with existing partner-based evaluations while testing whether PIP transfers across partially observable settings. Burrito-PO retains the Burrito task and partner protocol, whereas Overcooked-PO follows the partner-based evaluation structures used in GAMMA and GOAT while modifying the observation condition. This design limits confounding changes in task mechanics and partner construction, allowing the effect of intermittent partner visibility to be examined more directly. Burrito-PO provides the primary setting for partner-state inference, while Overcooked-PO tests whether the same approach remains effective under a different task structure and held-out-partner protocol. In both environments, explicit predictions of unobserved partner locations and actions allow us to assess partner-state inference accuracy alongside coordination performance.

OvercookedV2~\citep{gessler2025overcookedv2} also supports partial observations, but its original evaluation uses cross-play among independently trained agents and studies coordination under asymmetric task information and protocol formation. Adopting it would therefore change the coordination scenario and evaluation protocol in addition to the observation condition. We use Overcooked-PO to preserve continuity with existing partner-based evaluations, rather than because partner-state inference cannot be studied in OvercookedV2.

Melting Pot~\citep{agapiou2022melting} serves a complementary role by evaluating PIP under a different observation interface, task dynamics, and partner construction procedure. We therefore interpret its results as evidence of transfer across settings rather than directly comparing the magnitude of performance improvements with those observed in Burrito-PO and Overcooked-PO.

\section{Additional Analysis}



\subsection{Latent-Space Inspection via t-SNE}
\label{sec:supp_tsne}

\begin{figure*}[t]
    \centering
    \includegraphics[width=.99\linewidth]{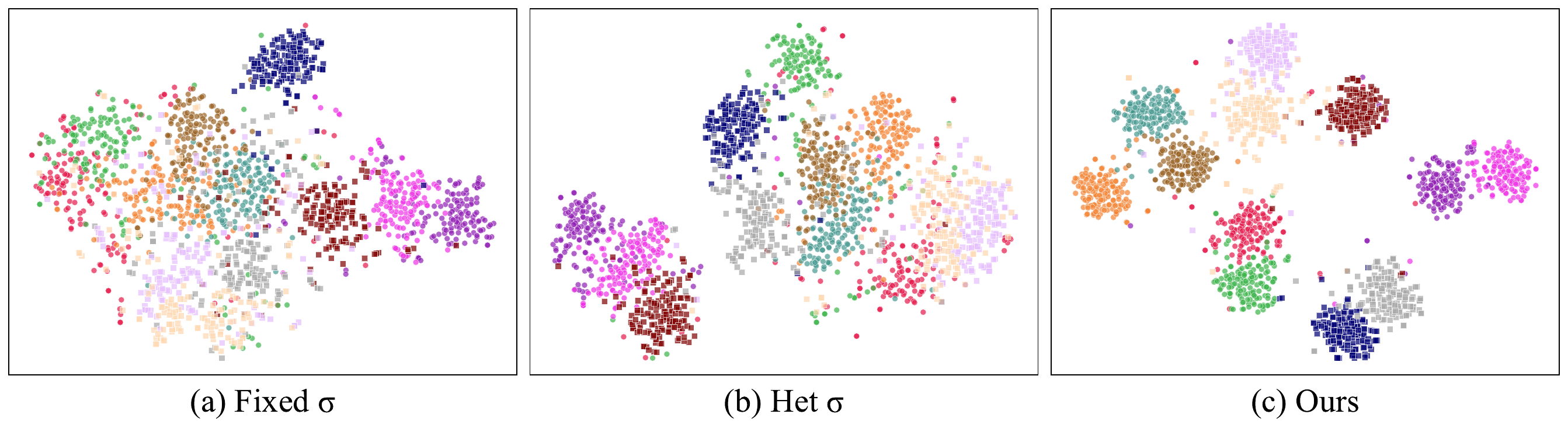}
    \caption{
        t-SNE visualization of episode-level posterior means from each encoder
        on held-out trajectories from 12 BP partners
        (144 episodes per partner; perplexity~50, PCA initialization).
        Each point corresponds to one episode and is colored by the partner
        policy that generated it, with the same color used for the same partner
        across panels. Compared with (a) Fixed~$\sigma$ and
        (b) Het~$\sigma$, (c) Ours produces more clearly
        separated partner-specific latent clusters.
    }
    \label{fig:tsne}
\end{figure*}

To complement the clustering metrics in the main paper, we qualitatively visualize the latent space of each encoder using t-SNE. For each held-out trajectory collected with one of the 12 held-out BP partners, we extract the episode-level posterior mean $\mu$, project the embeddings to two dimensions with t-SNE (perplexity 50, PCA initialization), and color each point by the BP partner policy that generated the trajectory. The same color denotes the same held-out BP partner across panels, and all encoders are evaluated on the same held-out BP trajectories under the same local $7\times7$ observation setting.

As shown in~\cref{fig:tsne}, the fixed-$\sigma$ baseline produces a broad latent region with heavily intermixed partner identities. The heteroscedastic baseline forms somewhat more structured regions, but many partners remain overlapped. In contrast, the KD-distilled student yields substantially more separated partner-specific clusters, even though it uses the same local observation input at test time as the baselines. This visualization is consistent with the higher NMI and ARI reported in the main paper, suggesting that posterior distillation transfers behaviorally meaningful partner information from the richer teacher view to the partial-observation student.




\subsection{Partner-state Belief Prediction Analyses}
\label{sec:supp_moved_belief_prediction}

\begin{figure*}[t]
    \centering
    \includegraphics[width=0.323\linewidth]{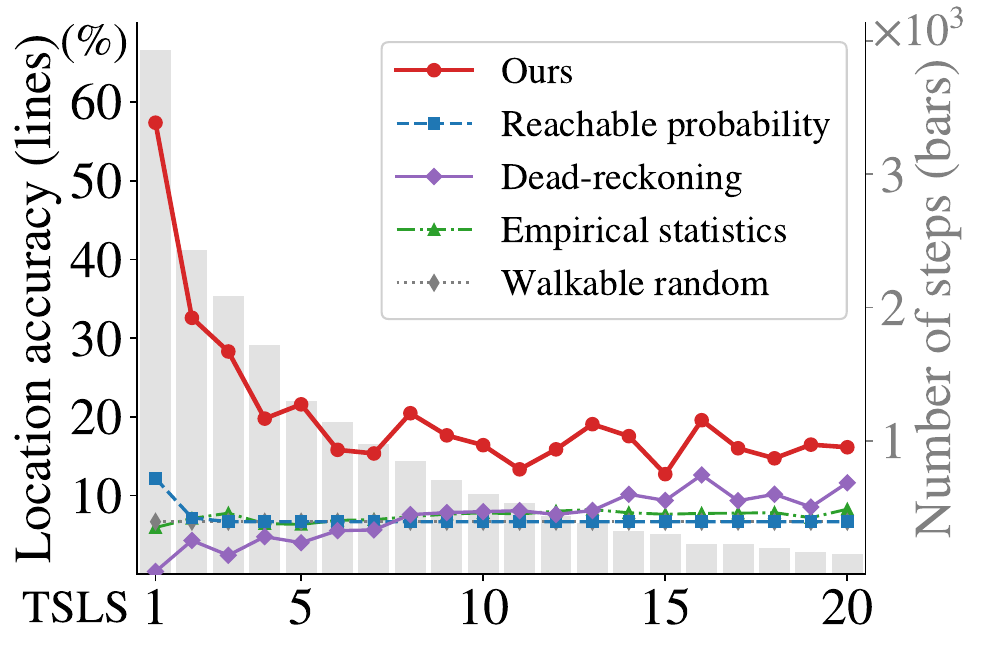}
    \hfill
    \includegraphics[width=0.323\linewidth]{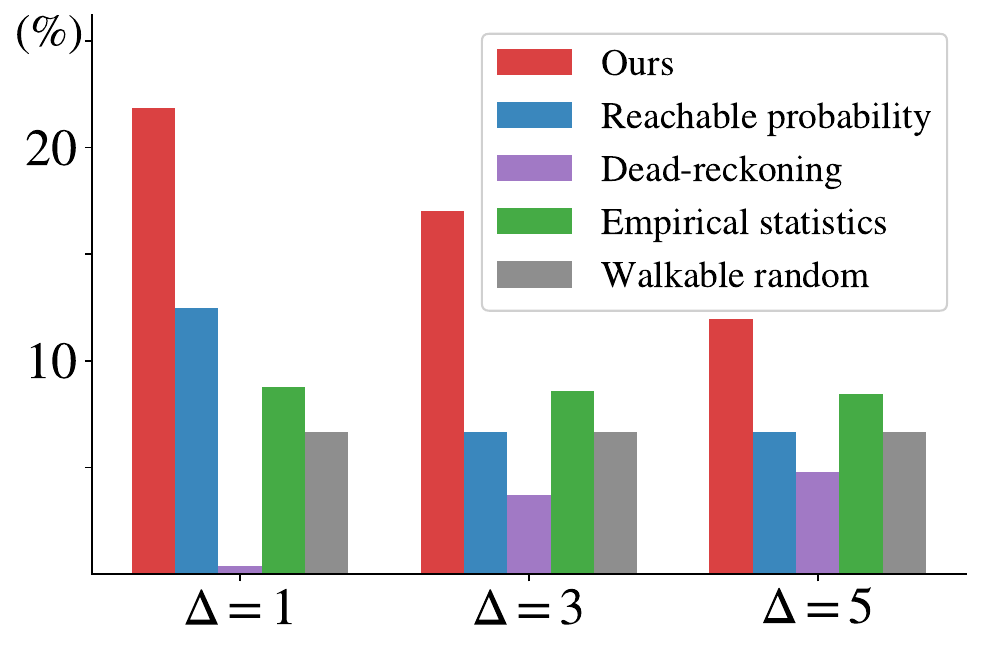}
    \hfill
    \includegraphics[width=0.323\linewidth]{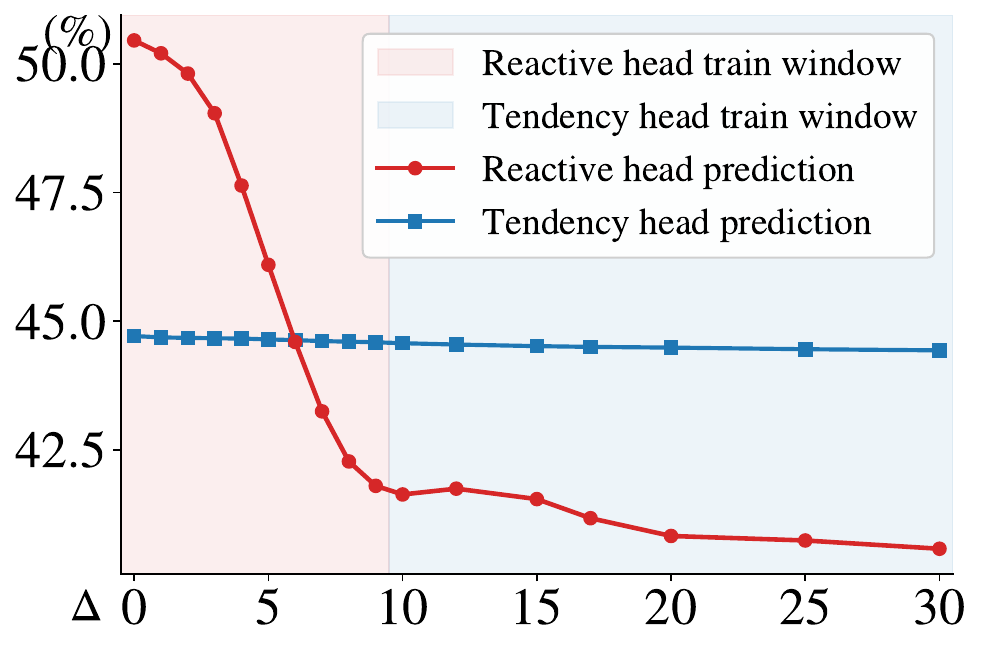}
    \caption{Partner-state Belief analysis. (a) Current Location Belief accuracy by Time Since Last Seen (TSLS). (b) $\Delta$-step future Location Belief accuracy. (c) $\Delta$-step future Behavioral Belief accuracy.}
    \label{fig:belief_quality}
\end{figure*}

We analyze whether the proposed belief-aware policy forms meaningful Partner-state Belief, as shown in~\cref{fig:belief_quality}. The belief branch is designed to capture two complementary aspects of the partner: where the partner is or will be, and what the partner is likely to do.

We first evaluate the Location Belief. Since exact cell-level localization under partial observability is unnecessarily strict, the model predicts a coarse location bin, where each bin corresponds to a $2 \times 2$ region of the map. In panel (a) of~\cref{fig:belief_quality}, we measure current coarse location prediction accuracy when the partner is outside the ego agent's field of view, grouped by Time Since Last Seen (TSLS), the number of steps since the partner left the visible range. To focus on practically meaningful cases, we evaluate prediction when the partner is moving. The proposed Location Belief consistently outperforms statistical baselines, including (i) probability spreading over time-dependent reachable regions (Reachable probability), (ii) movement-direction-based extrapolation (Dead-reckoning), (iii) empirical action statistics from the data (Empirical statistics), and (iv) random selection from the reachable region (Walkable random). The gap is especially clear in the small-TSLS region, where the partner has just disappeared from view. This indicates that our model forms an informative belief about the likely region of the currently invisible partner, rather than relying only on simple reachability statistics. Moreover, panel (b) shows that the model predicts the partner's $\Delta$-step future location more accurately than the baselines, suggesting that the Location Belief captures both current hidden location and near-future movement.

We next evaluate the Behavioral Belief, which consists of two temporal components: the reactive belief for near-future behavior and the tendency belief for broader action tendency. As shown in panel (c) of~\cref{fig:belief_quality}, the reactive belief achieves high accuracy at short horizons and gradually degrades as the prediction horizon increases, matching its role of capturing near-future actions. In contrast, the tendency belief remains more stable across timesteps, suggesting that it captures the partner's overall action distribution rather than a specific future action.
Taken together, these results show that the belief branch forms Partner-state Belief as intended: the Location Belief captures where the partner is and will be, while the Behavioral Belief captures both near-future actions and longer-term tendencies.

\subsection{Full-Observation Evaluation}
\label{sec:supp_moved_full_observation}


\begin{table}[t]
\centering
\small
\begin{tabular}{lc}
\toprule
Method & Score \\
\midrule
GAMMA& 143.3 $\pm$  11.2 \\
GOAT & 103.0 $\pm$  8.6 \\
PIP & \textbf{256.2 $\pm$  15.1 } \\
\bottomrule
\end{tabular}
\caption{Full-observation coordination performance on the Burrito Open layout with 12 held-out behavior-preference partners.}
\label{tab:fo_burrito_open}
\end{table}

To assess how closely our partial observability training pipeline approaches the performance attainable under full observability, we conduct a controlled experiment in which the entire pipeline is instead trained in a fully observable setting. Comparing the resulting performance with that of our proposed method allows us to quantify the performance gap caused by partial observability and evaluate how effectively our approach mitigates it. Specifically, partner dataset collection, VAE training, and target policy training are all performed under full observability. During evaluation, the fully observable-trained target policy acts with full observations, while the held-out BP partners receive only partial observations, identical to those used in the main experiments. We deliberately evaluate against the same held-out BP partners as in the main partial observability experiments, since BP agents differ substantially in task proficiency depending on their behavior preference. Reusing the identical evaluation partners eliminates partner-specific performance differences as a confounding factor, allowing the comparison to isolate the effect of the full-observability setting. The results are summarized in Table~\ref{tab:fo_burrito_open}.

The proposed method exhibits no significant difference between training under full observability and partial observability, showing only a marginal improvement of +7.2. This result suggests that, in this setting, the additional information available under FO does not translate into a measurable performance gain. In contrast, both baselines perform worse despite having access to the informational advantage of full observability during training. GAMMA shows a decrease of $-20.00$, GOAT shows a decrease of $-62.47$.



Two distinct factors are involved in this experiment: the information available to the coordinator itself, and the behavior distribution of the partners it is trained with. Regarding the former, the Open layout is the least occluded of the four maps, so the information removed by the PO setting is comparatively limited. This may partially account for the small difference observed for our method.
The latter factor, however, operates independently of how often the coordinator sees its partner.
Partners acting under partial observability may search or wait when uncertain about the other agent, and thus follow a behavior distribution that differs from that of fully observable partners.
An FO-trained coordinator is never exposed to such behaviors during training, so pairing it with PO evaluation partners introduces a partner-side distribution shift even when the coordinator's own observations are complete.
We hypothesize that GOAT's pronounced degradation primarily reflects this shift, while identifying why the adversarially trained generator is particularly sensitive to it is left for future work.

\subsection{Capacity and Supervision Controls}
\label{sec:supp_moved_controls}

\begin{table}[t]
\centering
\small
\setlength{\tabcolsep}{1.2mm}
\begin{tabular}{lccc}
\toprule
Method & Cap. & Sup. & Reward \\
\midrule
GAMMA & No & No & 163.3 $\pm$ 10.6 \\
+ Capacity control & Yes & No & 145.8 $\pm$ 5.7 \\
+ Supervision control & No & Yes & 88.7 $\pm$ 10.2 \\
+ Joint control & Yes & Yes & 96.0 $\pm$ 7.1 \\
\midrule
PIP (Ours) & Yes & Yes & \textbf{249.0 $\pm$ 12.2} \\
\bottomrule
\end{tabular}
\caption{Capacity and supervision controls on Burrito-PO Open. Cap.\ denotes capacity approximately matched to PIP, and Sup.\ denotes direct auxiliary supervision using privileged partner-state labels.}
\label{tab:matched_baselines}
\end{table}


To evaluate whether increased policy capacity or simple auxiliary partner-state prediction can reproduce PIP's gains, we construct three controls based on GAMMA. The capacity control enlarges GAMMA's policy GRU to approximately match PIP's trainable parameter count. The supervision control attaches cross-entropy heads to GAMMA's shared policy representation to predict future coarse partner locations and the partner's current action. The joint control combines both modifications. Unlike these controls, PIP uses dedicated belief modules that predict future locations and discounted future-action distributions and conditions the policy on their representations. Because their action targets differ, the supervision controls do not exactly match PIP's belief supervision. Results are reported in~\cref{tab:matched_baselines}.

None of the evaluated controls reproduces PIP's performance. The capacity control achieves a lower mean reward than GAMMA, while the supervision and joint controls achieve substantially lower mean rewards. These results indicate that approximately matching PIP's parameter count or adding the evaluated partner-state prediction heads to GAMMA is insufficient to account for PIP's gain under this setting.



\subsection{Robustness to Observation Corruption}

\begin{figure}[t]
    \centering
    \includegraphics[width=0.75\linewidth]{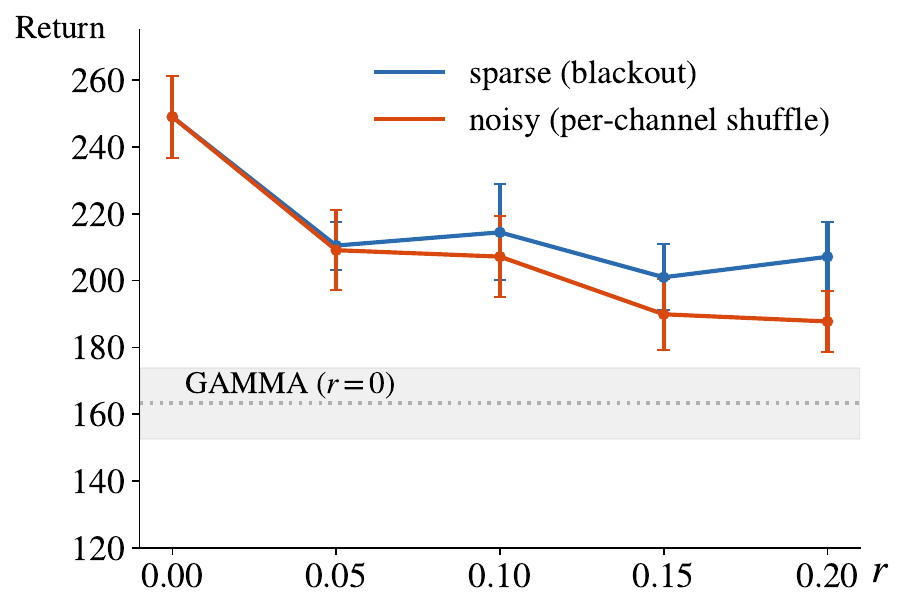}
    \caption{Return of PIP against the $M{=}12$ partners as the corruption rate $r$ increases from $0$ to $0.2$ (mean $\pm$ s.e.), under sparse (blackout) and noisy (per-channel shuffle) observation corruption. The dashed line is GAMMA with clean observations ($r=0$).}
    \label{fig:robustness}
\end{figure}


 We test how our method handles degraded observations by evaluating the converged PIP coordinator without further training in Burrito-PO Open layout. At test time, we corrupt its observation using one of two stochastic processes. The sparse process is a blackout in which the entire observation is masked with probability $r$ at each timestep, modelling intermittent sensor loss. The noisy process is a per-channel shuffle in which each observation channel is independently corrupted with probability $r$, and the values of each corrupted channel are spatially permuted within the field of view. This preserves the per-channel value distribution while destroying spatial correspondence, modelling per-modality scrambling. Thus, $r$ denotes the probability of masking an entire observation under sparse blackout and the independent probability of shuffling each channel under noisy corruption. We sweep $r$ from $0$ to $0.2$ and report the mean return against the held-out pool of $M{=}12$ partners (Mean $\pm$ S.E.;~\cref{fig:robustness}). PIP degrades gradually under both corruption processes. At $r{=}0.2$, the entire observation is masked at $20\%$ of timesteps under sparse blackout, whereas $20\%$ of observation channels are shuffled on average under noisy corruption. PIP retains returns of $207.1$ and $187.8$ under these respective conditions. Across the full range of $r$, PIP remains above GAMMA under the uncorrupted observation condition ($163.3 \pm 10.6$, dashed line), showing that its performance advantage persists under substantial test-time observation corruption.

\section{Human Evaluation}

\begin{figure*}[t]
\centering
\includegraphics[width=.55\linewidth]{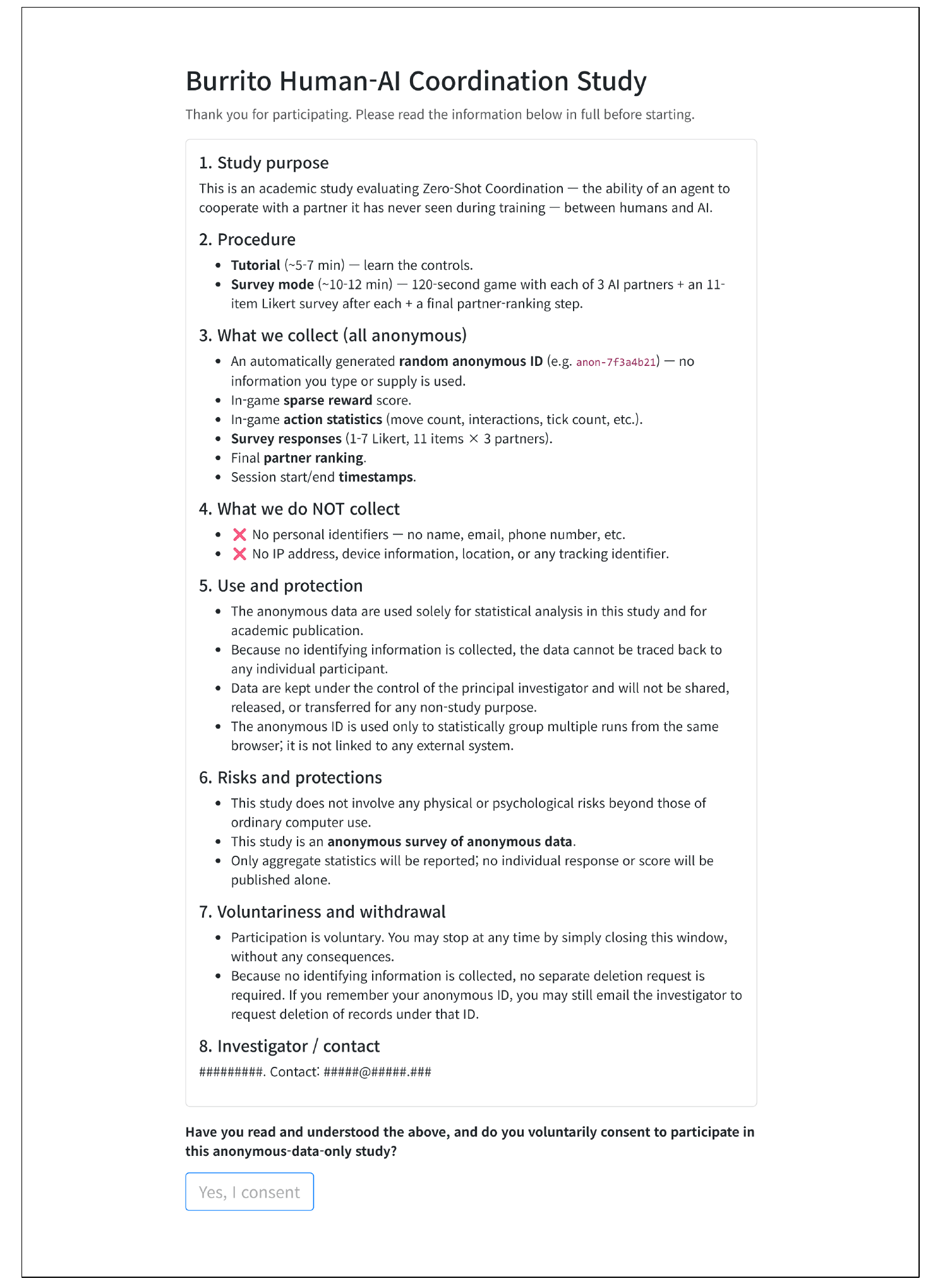}
\caption{Participant consent form for the Burrito human-AI coordination study, outlining the study purpose, procedure, anonymous data collection, exclusion of personal identifiers, data protection policy, potential risks, voluntary participation, and withdrawal conditions.}
\label{fig:participant_consent}
\end{figure*}

\subsection{Participant Consent and Data Protection}
\label{subsec:participant_consent}

As mentioned in the main paper, we voluntarily recruited participants for the experiment and provided them with sufficient time to understand the study procedure, potential risks, and data collection policy before obtaining their consent. As shown in~\cref{fig:participant_consent}, the consent form described the purpose of the human-AI coordination study, the overall experimental procedure, the types of anonymous data collected, and the purpose of data collection. The form also clarified that no personal identifiers, IP addresses, device information, location information, or other tracking identifiers were collected. To minimize privacy risks, each participant was assigned a random anonymous ID, and all collected data were used only for statistical analysis and research reporting. The study involved no physical or psychological risks beyond ordinary computer use, and participants were informed that their participation was voluntary and that they could withdraw from the study at any time.

\begin{table*}[t]
\centering
{\small
\begin{tabular}{p{0.05\linewidth} p{0.25\linewidth} p{0.6\linewidth}}
\toprule
\# & Step & Instruction \\
\midrule
1 & Movement & Use the arrow keys to move in all four directions. Participants were informed that the game only resumes after pressing Enter or the yellow ``Continue'' button. \\
2 & Read menu & Inspect the order cards at the bottom of the screen. Each card indicates the requested burrito type and the remaining ticks before the order expires. Orders can be delivered in any sequence. \\
3 & Pick meat & Move to the meat dispenser and press Space to pick up a piece of meat. \\
4 & Chop meat & Place the meat on the cutting board, press Space repeatedly to chop it, and pick up the resulting chopped meat. \\
5 & Grill meat & Place the chopped meat on the grill to cook it. \\
6 & Boil rice & Pick up rice from the rice dispenser and place it in the pot to boil. \\
7 & Assemble steak burrito & Pick up a clean plate and stack a tortilla, cooked meat, and boiled rice on the plate in any order. Participants were informed that ingredients can be added directly to a held plate using Space. \\
8 & Serve steak burrito & Carry the completed steak burrito to the delivery location and press Space to deliver it for score points. \\
9 & Pick mushroom & Move to the mushroom dispenser and press Space to pick up a mushroom. \\
10 & Chop mushroom & Place the mushroom on the cutting board, press Space repeatedly to chop it, and pick up the chopped mushroom. \\
11 & Grill mushroom & Place the chopped mushroom on the grill to fry it. \\
12 & Boil rice again & Prepare another portion of rice by placing rice from the dispenser into the pot. \\
13 & Assemble mushroom burrito & Stack a tortilla, fried mushroom, and boiled rice on a clean plate in any order to make a mushroom burrito. \\
14 & Serve mushroom burrito & Carry the completed mushroom burrito to the delivery location and press Space to deliver it. \\
15 & Burn demonstration & Place rice in the pot and wait until it burns into charcoal, demonstrating the consequence of overcooking. \\
16 & Extinguish fire & Pick up the fire extinguisher, face the burning station, and press Space to put out the fire. \\
17 & Trash charcoal & Put down the fire extinguisher, pick up the charcoal, and dispose of it in the trash bin. \\
18 & Wash plate & Pick up a dirty plate, bring it to the sink, and press Space several times to wash it clean. \\
19 & Free play & After all instructions were completed, participants freely played the game for 60 seconds before proceeding to the main survey. \\
\bottomrule
\end{tabular}
}
\caption{Tutorial steps provided before the human-AI coordination survey. The tutorial familiarized participants with the game controls, object interactions, burrito recipes, and failure-recovery mechanics.}
\label{tab:tutorial_steps}
\end{table*}

\subsection{Gameplay Interface}
\label{subsec:gameplay_interface}

\begin{figure}[!t]
    \centering
    \includegraphics[width=0.445\linewidth]{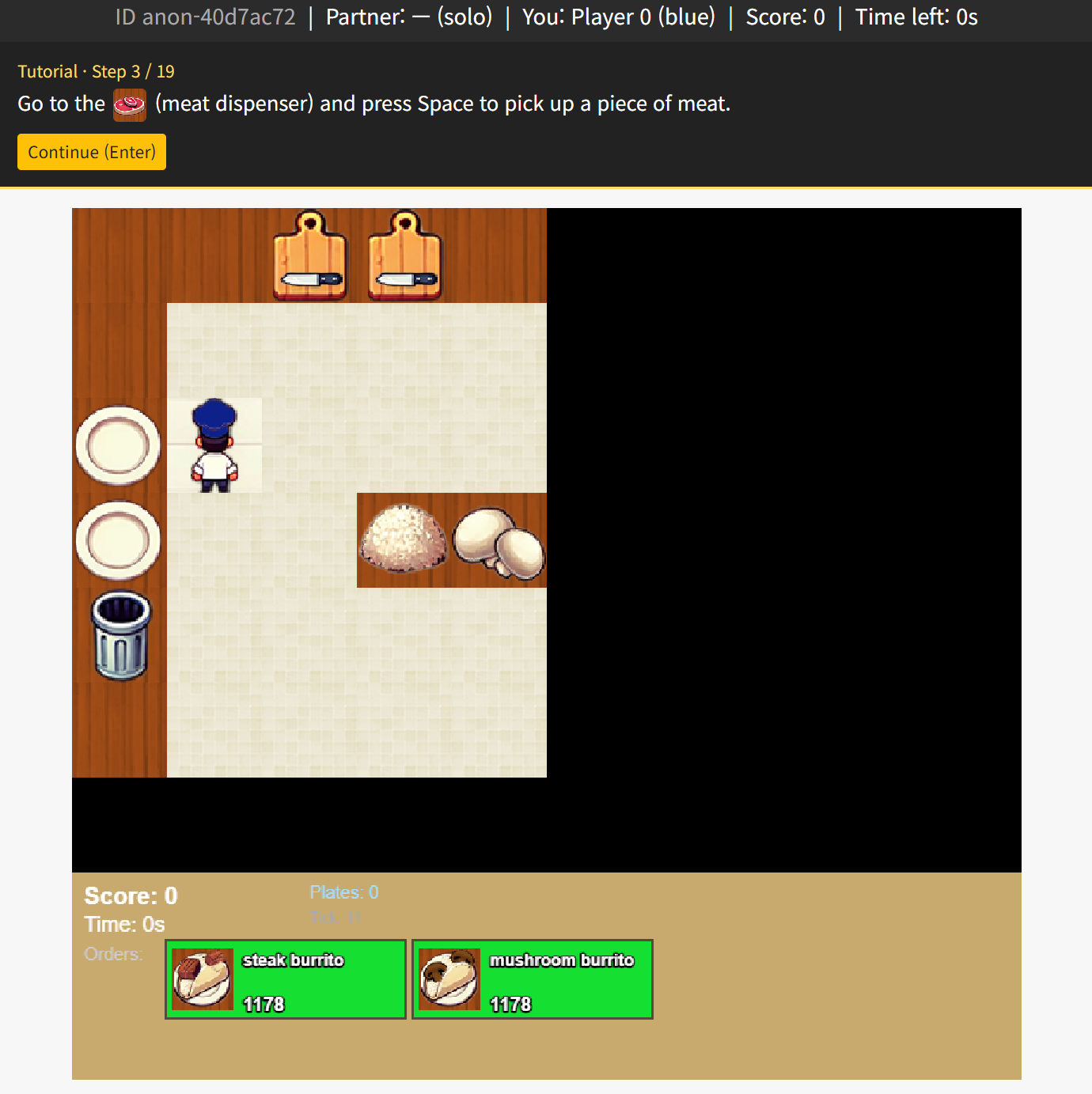}
    \hfill
    \includegraphics[width=0.50\linewidth]{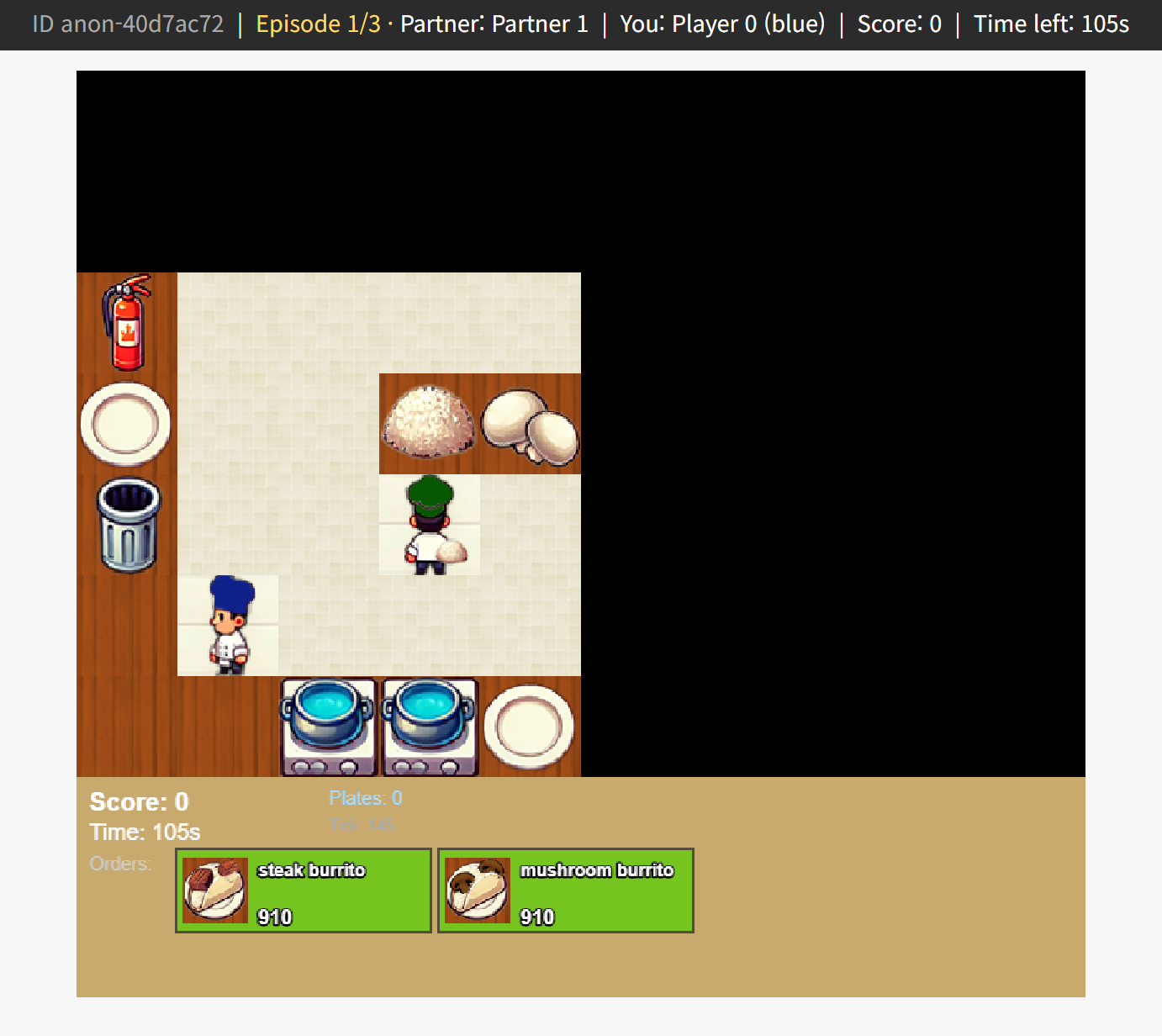}
    \caption{Tutorial and survey interfaces. The left panel shows step-by-step instructions, object icons, and the ego-centric partial-observation view. The right panel shows the main-survey gameplay interface, where participants cooperate with an anonymized AI partner under ego-centric partial observability.}
    \label{fig:gameplay_interfaces}
\end{figure}

Before the main user study, participants completed an interactive tutorial to familiarize themselves with the game rules, controls, object interactions, and burrito-making procedure. As shown in the left panel of~\cref{fig:gameplay_interfaces}, the tutorial interface displayed the current step, task instruction, relevant object icon, and a yellow ``Continue'' button. Participants advanced each step by clicking the button or pressing Enter, and the game timer was paused while instructions were displayed to avoid time pressure during reading. The tutorial guided participants through the essential gameplay mechanics, including movement, order-card interpretation, ingredient collection, chopping, cooking, plating, serving, fire handling, disposal of burned food, and dish washing, as summarized in~\cref{tab:tutorial_steps}. After completing the guided steps, participants were given a 60-second free-play period to further practice the controls and gameplay loop before the main survey, helping ensure a fairer comparison across AI partners.

After the tutorial, participants played the main survey episodes with AI partners using the gameplay interface shown in the right panel of~\cref{fig:gameplay_interfaces}. Each participant interacted with three different AI policies, GAMMA, GOAT, and PIP (Ours), in a randomized order, and played one 120-second (1200-timestep) episode with each policy. The interface displayed the anonymous participant ID, episode information, assigned AI partner, participant role and color, current score, remaining time, active order cards, plate count, and environment tick. The actual policy identity corresponding to the assigned AI partner, \textit{i.e.}, whether it was GAMMA, GOAT, or PIP (Ours), was not revealed to participants during gameplay. The survey gameplay followed an ego-centric partial-observation setting: participants could observe only a limited region centered on their own character, while areas outside the visible region were masked in black and updated as the participant moved. Thus, participants evaluated each anonymized AI partner under incomplete information about both the partner and the full task state.

\begin{figure*}[t]
\centering
\includegraphics[width=.99\linewidth]{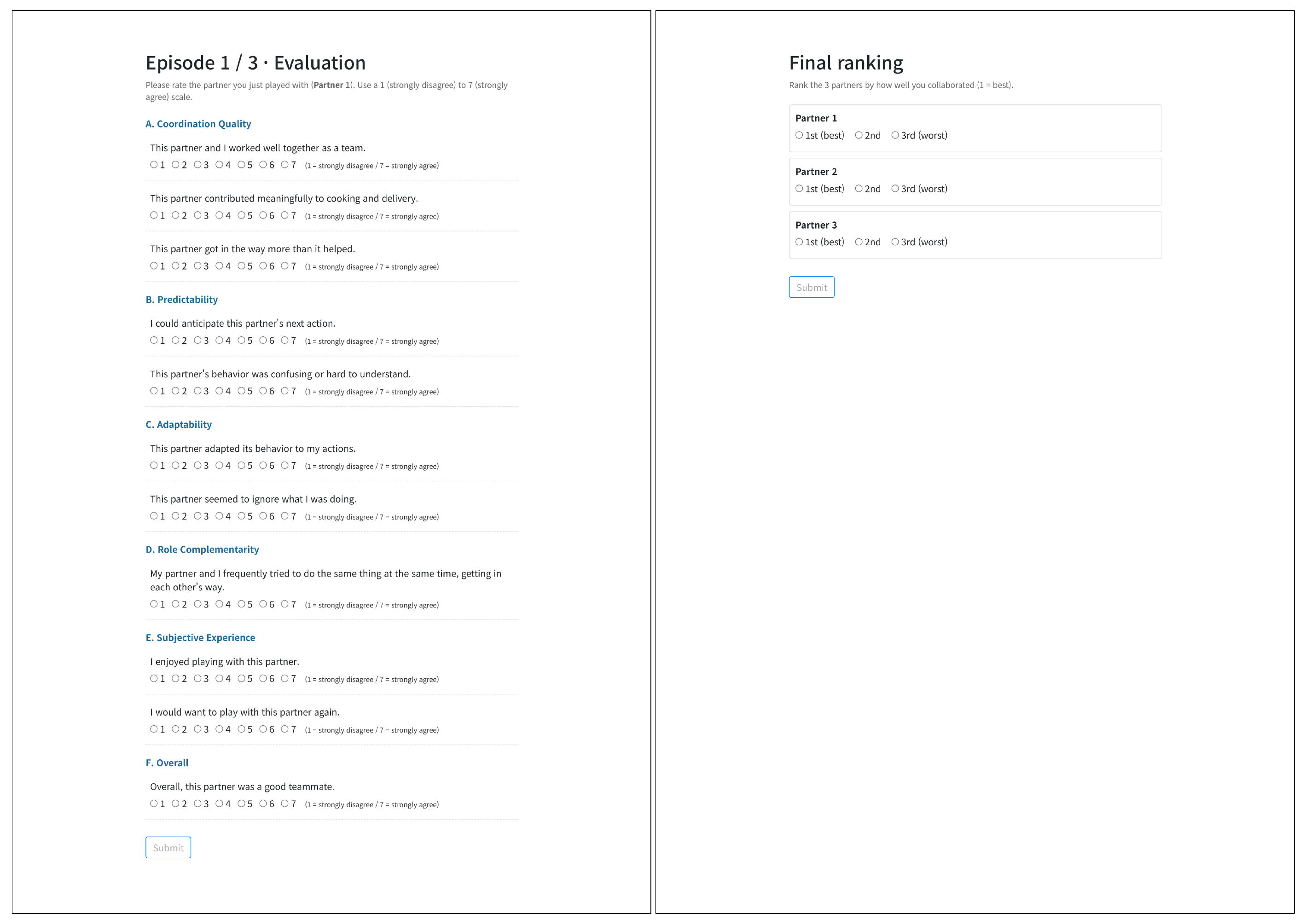}
\caption{Post-game evaluation materials. The left panel shows the post-episode evaluation survey assessing each AI partner after gameplay. The right panel shows the final preference ranking of the three AI partners based on collaboration experience.}
\label{fig:survey}
\end{figure*}

\begin{table*}[!t]
\centering
{\small
\begin{tabular}{lccccc}
\toprule
Metric & GAMMA & GOAT & PIP (Ours) & vs. GAMMA & vs. GOAT \\
\midrule
Sparse reward mean $\pm$ SE & $355.7 \pm 19.7$ & $370.4 \pm 18.4$ & $476.0 \pm 22.7$ & $+33.8\%$ & $+28.5\%$ \\
Top-1 preference & $10/66$, $15.2\%$ & $8/66$, $12.1\%$ & $48/66$, $72.7\%$ & $+57.5$ pts. & $+60.6$ pts. \\
\bottomrule
\end{tabular}
}
\caption{Human evaluation performance and preference summary ($N = 66$ participants).}
\label{tab:human_perf_pref}
\end{table*}

\subsection{Post-Game Evaluation and Preference Survey}
\label{subsec:post_game_survey}

After each gameplay episode, participants completed a post-episode evaluation survey to assess the AI partner they had just played with. As shown in the left panel of~\cref{fig:survey}, the survey used Likert-scale questions to measure participants' subjective evaluation of the agent, including its coordination quality, predictability, adaptability, role complementarity, subjective experience, and overall teamwork. After completing all gameplay episodes with the three AI partners, participants also provided a final preference ranking. As shown in the right panel of~\cref{fig:survey}, this ranking asked participants to order the three AI partners according to their overall collaboration experience.

\subsection{Human Evaluation Details}
\label{subsec:human_eval_details}

We provide additional quantitative analysis of the human evaluation results. Each participant interacted with all three anonymized agents, GAMMA, GOAT, and PIP (Ours), under the same ego-centric partial-observation setting. We therefore treat the human evaluation as a within-subject comparison and analyze task performance, preference rankings, and post-episode survey scores.

\paragraph{Survey Score Construction.}
Each post-episode survey item was rated on a 1--7 Likert scale. Before computing category-level scores, negatively worded items were reverse-coded so that higher values consistently indicate better perceived partner quality. Each category score is then computed as the mean of its constituent items after this transformation.

\paragraph{Task Performance and Preference Ranking.}
\Cref{tab:human_perf_pref} summarizes task-level sparse reward and final preference ranking. PIP achieves the highest sparse reward among the three agents, with a mean reward of 476.0 compared with 355.7 for GAMMA and 370.4 for GOAT. This corresponds to a 33.8\% improvement over GAMMA and a 28.5\% improvement over GOAT. PIP is also ranked as the most preferred partner by 48 out of 66 participants, corresponding to 72.7\% of all participants. In comparison, GAMMA and GOAT are selected as the top-ranked partner by 15.2\% and 12.1\% of participants, respectively. The full preference distribution is shown in panel (b) of the human-evaluation figure in the main paper.


\begin{table}[!t]
\centering
\setlength{\tabcolsep}{4pt}
{\small
\begin{tabular}{lcccccc}
\toprule
\multirow{2}{*}{Outcome} & \multicolumn{2}{c}{Friedman test} & \multicolumn{2}{c}{PIP vs. GAMMA} & \multicolumn{2}{c}{PIP vs. GOAT} \\
\cmidrule(lr){2-3} \cmidrule(lr){4-5} \cmidrule(lr){6-7}
 & $\chi^2(2)$ & $p$ & $p$ & $r$ & $p$ & $r$ \\
\midrule
Sparse reward   & 34.95 & $2.57{\times}10^{-8}$ & $2.43{\times}10^{-7}$ & $+0.74$ & $3.16{\times}10^{-6}$ & $+0.66$ \\
Preference rank & 37.85 & $6.04{\times}10^{-9}$ & $4.80{\times}10^{-6}$ & $+0.62$ & $1.16{\times}10^{-6}$ & $+0.67$ \\
\bottomrule
\end{tabular}
}
\caption{Within-subject statistical tests for task performance and preference ranking.}
\label{tab:human_main_stats}
\end{table}

\begin{table*}[!t]
\centering
{\small
\begin{tabular}{lccccc}
\toprule
Category & GAMMA & GOAT & PIP & vs. GAMMA & vs. GOAT \\
\midrule
Coordination & $3.68$ & $3.26$ & $5.01$ & $+36.1\%$ & $+53.7\%$ \\
Predictability & $3.33$ & $3.11$ & $4.83$ & $+45.0\%$ & $+55.3\%$ \\
Adaptability & $3.29$ & $3.11$ & $4.89$ & $+48.6\%$ & $+57.2\%$ \\
Role Complementarity & $3.09$ & $2.80$ & $4.35$ & $+40.8\%$ & $+55.4\%$ \\
Subjective Experience & $3.33$ & $3.13$ & $5.02$ & $+50.8\%$ & $+60.4\%$ \\
Overall & $3.45$ & $3.18$ & $5.27$ & $+52.8\%$ & $+65.7\%$ \\
\bottomrule
\end{tabular}
}
\caption{Category-level survey scores and relative improvements.}
\label{tab:human_survey_scores}
\end{table*}


\begin{table}[!t]
\centering
\setlength{\tabcolsep}{4pt}
{\small
\begin{tabular}{lcccccc}
\toprule
\multirow{2}{*}{Category} & \multicolumn{2}{c}{Friedman test} & \multicolumn{2}{c}{PIP vs. GAMMA} & \multicolumn{2}{c}{PIP vs. GOAT} \\
\cmidrule(lr){2-3} \cmidrule(lr){4-5} \cmidrule(lr){6-7}
 & $\chi^2(2)$ & $p$ & $p$ & $r$ & $p$ & $r$ \\
\midrule
Coordination          & 35.46 & $1.99{\times}10^{-8}$  & $2.14{\times}10^{-5}$ & $+0.62$ & $1.42{\times}10^{-7}$ & $+0.76$ \\
Predictability        & 39.07 & $3.28{\times}10^{-9}$  & $6.64{\times}10^{-7}$ & $+0.72$ & $4.31{\times}10^{-8}$ & $+0.82$ \\
Adaptability          & 33.95 & $4.25{\times}10^{-8}$  & $1.14{\times}10^{-5}$ & $+0.66$ & $1.23{\times}10^{-6}$ & $+0.72$ \\
Role Complementarity  & 28.31 & $7.12{\times}10^{-7}$  & $7.10{\times}10^{-5}$ & $+0.60$ & $3.01{\times}10^{-6}$ & $+0.73$ \\
Subjective Experience & 32.57 & $8.47{\times}10^{-8}$  & $1.20{\times}10^{-6}$ & $+0.73$ & $5.86{\times}10^{-7}$ & $+0.76$ \\
Overall               & 42.90 & $4.83{\times}10^{-10}$ & $2.70{\times}10^{-7}$ & $+0.79$ & $1.85{\times}10^{-7}$ & $+0.78$ \\
\bottomrule
\end{tabular}
}
\caption{Supporting within-subject statistical tests for post-episode survey scores. Wilcoxon $p$-values are uncorrected, and all PIP-vs-baseline comparisons remain significant after Holm--Bonferroni correction.}
\label{tab:human_survey_stats}
\end{table}

\paragraph{Statistical Analysis.}
Because each participant evaluated all three agents, we use within-subject nonparametric tests and report p-values rather than relying on the visual overlap of standard-error bars. The significance markers shown in panel (a) of the human-evaluation figure in the main paper correspond to the paired Wilcoxon signed-rank tests reported here. We first apply Friedman tests~\citep{friedman1937use} across the three agents, followed by paired Wilcoxon signed-rank tests~\citep{wilcoxon1992individual} comparing PIP against each baseline, and report matched-pairs rank-biserial correlation $r$ as the effect size. As summarized in \cref{tab:human_main_stats}, the Friedman tests show significant differences across agents for both sparse reward and preference rank. The paired Wilcoxon tests show that PIP significantly outperforms both GAMMA and GOAT on both metrics, with large effect sizes ($r \geq 0.62$). Reported paired Wilcoxon signed-rank test values are uncorrected p-values, and all paired comparisons remain significant after Holm-Bonferroni correction across the four PIP-vs-baseline tests in \cref{tab:human_main_stats}. For preference rank, the sign of $r$ is oriented so that $r>0$ indicates that PIP achieves a lower, better rank than the baseline.

\paragraph{Survey Score Decomposition.}

\Cref{tab:human_survey_scores} reports the category-level scores and relative improvements of human evaluation. PIP receives the highest score in every category. The largest relative gains appear in Overall, Subjective Experience, Predictability, and Adaptability, suggesting that participants perceived PIP not only as more task-effective, but also as easier to understand, more responsive to their behavior, and more desirable to cooperate with under partial observability.

As a supporting analysis, we also test each survey category using within-subject nonparametric tests and report matched-pairs rank-biserial correlation~\citep{kerby2014simple} $r$ as the effect size, where $|r| \geq 0.5$ is conventionally considered a large effect. As summarized in~\cref{tab:human_survey_stats}, Friedman tests indicate significant differences among the three agents for all six survey categories. Paired Wilcoxon signed-rank tests show that PIP significantly outperforms both baselines in every category, with large effect sizes throughout ($r \geq 0.60$). Reported paired Wilcoxon p-values are uncorrected, and these paired survey comparisons remain significant after Holm-Bonferroni correction~\citep{holm1979simple} across the 12 PIP-vs-baseline survey comparisons in \cref{tab:human_survey_stats}.

Overall, the human evaluation suggests that the proposed agent improves both objective and subjective aspects of human-AI coordination. Participants achieved higher sparse reward with PIP, selected it as the most preferred partner substantially more often, and gave it higher ratings across all survey dimensions. These results indicate that the proposed framework improves not only task-level coordination, but also perceived predictability, adaptability, role complementarity, and overall cooperation quality under ego-centric partial observability.